\documentclass[10pt,twocolumn,letterpaper]{article}

\usepackage{iccv}
\usepackage{times}
\usepackage{epsfig}
\usepackage{graphicx}
\usepackage{amsmath}
\usepackage{amssymb}

\usepackage[pagebackref=true,breaklinks=true,colorlinks,bookmarks=false]{hyperref}

\iccvfinalcopy %

\def\iccvPaperID{11105} %

\ificcvfinal\fi

\hypersetup{
    colorlinks=true,
    linkcolor=blue,
    filecolor=magenta,      
    urlcolor=cyan,
    }
    
\usepackage{color,xcolor}
\usepackage{epsfig}
\usepackage{graphicx}

\usepackage{adjustbox}
\usepackage{array}
\usepackage{booktabs}
\usepackage{colortbl}
\usepackage{float,wrapfig}
\usepackage{hhline}
\usepackage{multirow}
\usepackage{caption}

\usepackage{cleveref}

\usepackage{bm}
\usepackage{nicefrac}
\usepackage{mathtools}

\usepackage{changepage}
\usepackage{extramarks}
\usepackage{fancyhdr}
\usepackage{lastpage}
\usepackage{setspace}
\usepackage{xspace}

\usepackage{url}

\usepackage{amssymb}
\usepackage{amsfonts}
\usepackage{algorithmic}
\usepackage{algorithm}
\usepackage{enumitem}
\usepackage{bbm}
\usepackage[super]{nth}

\hypersetup{colorlinks=true,linkcolor=red,citecolor=brown,urlcolor=blue}

\makeatletter
\DeclareRobustCommand\onedot{\futurelet\@let@token\@onedot}
\def\@onedot{\ifx\@let@token.\else.\null\fi\xspace}

\def\eg{\emph{e.g}\onedot} 
\def\ie{\emph{i.e}\onedot}

\makeatother

\renewcommand{\paragraph}[1]{\vspace{.1em}\noindent\textbf{#1}}

\newcommand{\sect}[1]{Section~\ref{#1}}

\newcommand{\eqn}[1]{Equation~(\ref{#1})}
\newcommand{\fig}[1]{Figure~\ref{#1}}

\newcommand{\ignore}[1]{}

\definecolor{rowblue}{RGB}{220,230,240}
\definecolor{myorchid}{RGB}{150,10,30}
\definecolor{myblue}{RGB}{10,30,250}
\definecolor{mygreen}{RGB}{10,120,10}

\usepackage{amsmath,amsfonts,bm}

\def\eqref#1{equation~\ref{#1}}

\def\1{\bm{1}}

\def\vc{{\bm{c}}}

\def\vw{{\bm{w}}}
\def\vx{{\bm{x}}}

\DeclareMathAlphabet{\mathsfit}{\encodingdefault}{\sfdefault}{m}{sl}
\SetMathAlphabet{\mathsfit}{bold}{\encodingdefault}{\sfdefault}{bx}{n}

\newcommand{\R}{\mathbb{R}}

\begin{document}

\title{Unsupervised Compositional Concepts Discovery \\ with Text-to-Image Generative Models}

\author{
Nan Liu$^{1}$$^*$$\;\;\;$
Yilun Du$^{2}$$^*$ $\;\;\;$ 
Shuang Li$^{2}$$^*$ $\;\;\;$ 
Joshua B. Tenenbaum$^{2}$ $\;\;$ 
Antonio Torralba$^{2}$ $\;\;$  \\
$^1$UIUC\;\;\;  $^2$MIT\;\;\;   \\
* indicates equal contribution \\
Website: \href{https://energy-based-model.github.io/unsupervised-concept-discovery/}{{https://energy-based-model.github.io/unsupervised-concept-discovery/}} \\
}
 
\twocolumn[{%
\renewcommand\twocolumn[1][]{#1}%
\maketitle
\centering
\includegraphics[width=1\linewidth]{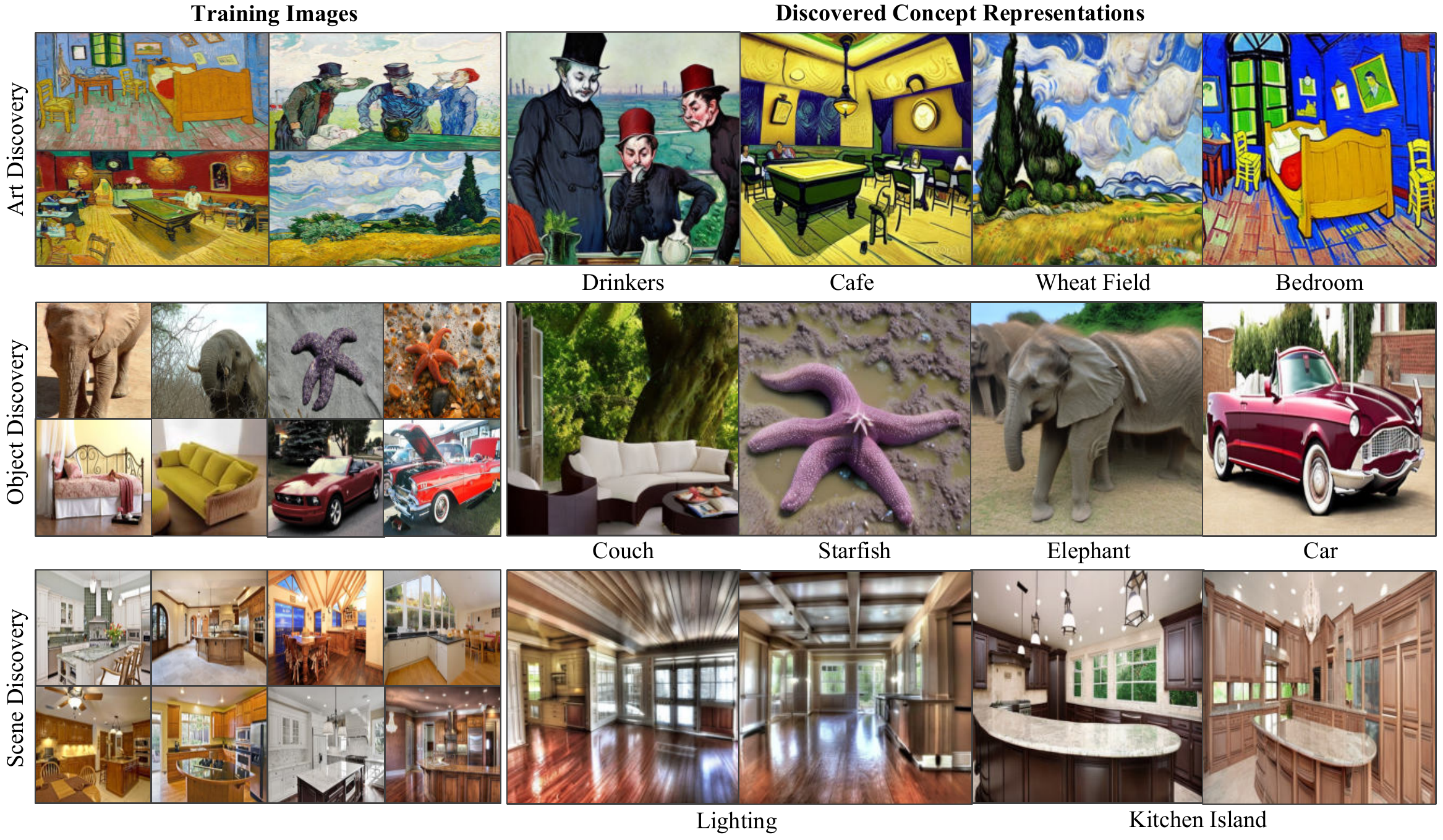}
\vspace{-20pt}
\captionof{figure}{\small{\textbf{Unsupervised Image Decomposition.}
       Our approach is able to decompose a dataset of unlabeled images into different concepts. 
       We name each decomposed concept for easy understanding.
    }}
\label{fig:teaser}
\vspace{15pt}

}]

\maketitle

\ificcvfinal\thispagestyle{empty}\fi

\begin{abstract}
Text-to-image generative models have enabled high-resolution image synthesis across different domains, but require users to specify the content they wish to generate. In this paper, we consider the inverse problem -- given a collection of different images, can we discover the generative concepts that represent each image? We present an unsupervised approach to discover generative concepts from a collection of images, disentangling different art styles in paintings, objects, and lighting from kitchen scenes, and discovering image classes given ImageNet images. We show how such generative concepts can accurately represent the content of images, be recombined and composed to generate new artistic and hybrid images, and be further used as a representation for downstream classification tasks.

\end{abstract}
\section{Introduction}

When presented with a set of images, we can infer and discover common concepts across images. For instance, given a set of images of kitchen scenes in \fig{fig:teaser}, we can grasp different illumination patterns in the kitchen and identify various elements within kitchens, such as dining tables, kitchen islands, and cabinets. Moreover, we possess the ability to conjure up vivid mental images of new scenes that combine elements between different kitchen scenes or visualize how these elements may manifest in unfamiliar settings -- envisioning, for instance, how a dining table may appear in a forest.

Can we construct computer vision systems that may likewise understand, recombine, and imagine the visual world?  Most existing work in concept discovery focus on discovering latent vectors or directions representing individual concepts~\cite{gal2022image,jahanian2019steerability,harkonen2020ganspace, shen2020interpreting,wu2021stylespace}, but require supervised data labeling each concept.  Other works have focused on discovering compositional generative concepts from images but focus only on discovering objects~\cite{burgess2019monet,locatello2020objectcentric}.  Recently, COMET~\cite{du2021comet} proposes an approach to decompose scenes into a set of generative concepts representing both global scene concepts, such as lighting and camera position, and local concepts, such as objects. However, the approach is only applied to simple datasets and fails to generate complex photorealistic images.

In this work, we illustrate how we can leverage the rich semantic information in large text-to-image generative models to discover a set of diverse compositional generative concepts from unlabeled natural images. Our work extends the approach in~\cite{du2021comet} using the interpretation of diffusion models as EBMs~\cite{liu2022compositional} and decomposes each image into a set of different probability distributions. We illustrate how each decomposed probability distribution captures different global and local scene concepts in an image, ranging from ImageNet class identity to portions of images such as islands and cabinets in a kitchen.

In \fig{fig:teaser}, we show how our approach can discover compositional concepts across a wide set of different domains. In the top row of \fig{fig:teaser}, we illustrate how our approach can discover different art concepts, such as wheat fields, cafes, and bedrooms, from paintings by either Van Gogh or Claude Monet. In the middle row of \fig{fig:teaser}, we demonstrate how our approach can discover classes of images, such as couches, starfish, elephants, and cars, from a collection of unlabeled ImageNet images. Finally, in the bottom row of \fig{fig:teaser}, we show how our approach can discover the compositional components of a kitchen, such as lighting patterns and kitchen islands.

In this work, we contribute the following: \textbf{(1)} We illustrate a scalable approach to discover unsupervised compositional concepts in realistic images using existing generative models. \textbf{(2)} Our method achieves state-of-the-art performance on concept discovery across different domains, in both global and local concept discovery, such as automatically discovering painting styles, and decomposing scenes into lighting and objects. \textbf{(3)} We illustrate that the discovered generative concepts can be used for diverse tasks, such as generating novel creative images or as effective representations for downstream classification tasks.

\section{Related Works}

\paragraph{Compositional Generation.} Compositional generation, where we seek to generate an image subject to a set of underlying specifications, has attracted increasing attention in recent years~\cite{du2020compositional,liu2021learning,liu2022compositional,feng2022training,shicompositional,cong2023attribute,cho2023,du2023reduce,huang2023composer,lace, wang2023concept,li2022composing,wang2023compositional,sohn2023learning,huang2023collaborative}. Existing work on compositional generation focuses either on modifying the feedforward generative process to focus on a set of specifications~\cite{feng2022training,shicompositional,cong2023attribute,huang2023composer,huang2023collaborative}, or by composing a set of independent models specifying desired constraints~\cite{du2020compositional, liu2021learning,  liu2022compositional,lace,du2023reduce,wang2023concept,li2022composing}. Our work utilizes the compositional operators defined from ~\cite{du2020compositional, liu2022compositional}, but aims to discover a set of compositional components from an unlabeled dataset of images.

\paragraph{Unsupervised Concept Discovery.} Existing works in concept discovery in computer vision typically focus on discovering a latent space to manipulate images~\cite{gal2022image,jahanian2019steerability,harkonen2020ganspace, shen2020interpreting,wu2021stylespace,ruiz2022dreambooth} but require supervised data to specify each concept. Some work has focused instead on discovering multiple concepts from images,  but focus on discovering objects represented as separate segmentation masks~\cite{burgess2019monet,locatello2020objectcentric, du2021unsupervised}. Most similar to our work is that of COMET~\cite{du2021comet}, which decomposes images into a set of composable energy functions representing both objects and scene-level factors such as lighting or camera position. Our work builds on this work, but represents each individual energy function with a diffusion model. We illustrate how this enables us to generate and decompose complex, high-resolution images. 

\paragraph{Text-Conditioned Generative Modeling.}
In recent years, tremendous efforts have been made towards text-based 2D and 3D synthesis using various types of generative models, including GANs~\cite{goodfellow2020generative}, VAEs~\cite{kingma2013auto}, Normalizing Flows~\cite{rezende2015variational}, Energy-Based Models~\cite{lecun2006tutorial,du2019implicit} and Diffusion Models~\cite{sohl2015deep,ho2020denoising}. Diffusion models have become the de facto method for 2D text-to-image synthesis~\cite{nichol2021glide,saharia2022photorealistic,ramesh2022hierarchical,ruiz2022dreambooth,liew2022magicmix,chefer2023attend,brooks2022instructpix2pix,brack2023sega,wu2022uncovering,brack2022stable,gokhale2022benchmarking,rombach2022high,gal2022image,kumari2023multi} and text-to-3D synthesis~\cite{poole2022dreamfusion,lin2022magic3d}. Most relevant to our work, textual inversion~\cite{gal2022image} leverages pre-trained text-to-image diffusion models to map a visual concept to a single-word representation (\ie, a supervised approach). In contrast, we demonstrate how such diffusion models can be leveraged to discover multiple visual representations from a set of images simultaneously without using image labels.

\section{Background}
\begin{figure*}[t]
\begin{center}
\includegraphics[width=\textwidth]{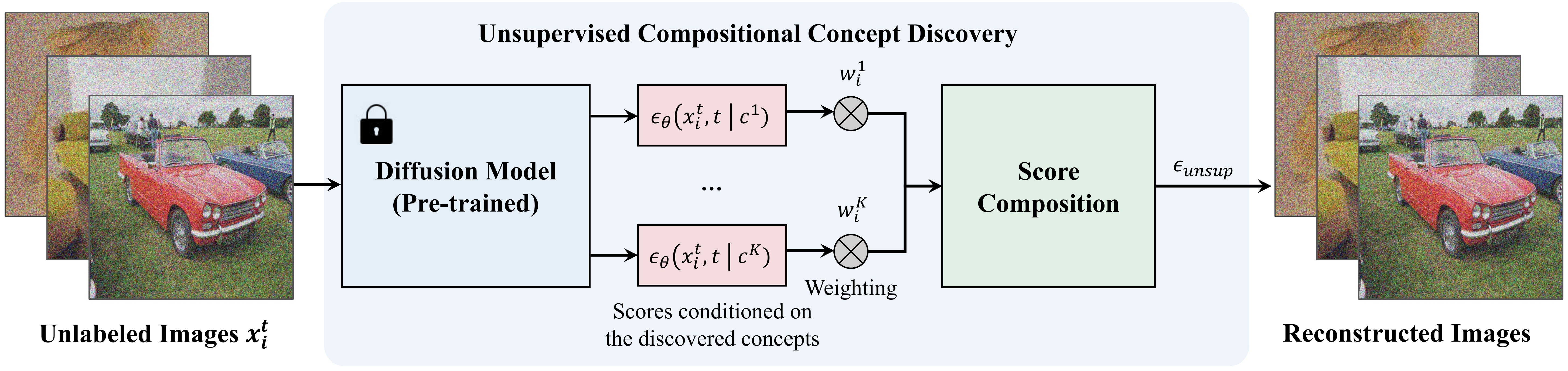}
\end{center}
\vspace{-15pt}
\caption{\small \textbf{Compositional Concept Discovery.} We discover a set of compositional concepts given a dataset of unlabeled images. Score functions representing each concept $\{\vc^1, \dots, \vc^K\}$ are composed together to form a score function $\epsilon_{\text{unsup}}$ that is trained to denoise images. The inferred concepts can be used to generate new images.
}
\label{fig:model}
\end{figure*}

In this section, we introduce background knowledge on diffusion models and on composing different concepts with diffusion models.

\subsection{Denoising Diffusion Probabilistic Models}

Denoising Diffusion Probabilistic Models (DDPMs)~\cite{sohl2015deep,ho2020denoising} are a class of generative models where the generation of images $\vx_0$ is formed by iteratively denoising an image corrupted with Gaussian noise. Given a randomly sampled noise $\epsilon \sim \mathcal{N}(0, 1)$, and a set of $t$ different noise levels $\epsilon^t = \alpha^t \epsilon$ added to a clean image $\vx_0$\footnote{Note that in practice, $x_0+\epsilon^t$ is also scaled by a contraction $\beta^t$ before being fed into the diffusion model.}, a denoising model $\epsilon_\theta$ is trained to denoise the image at specified noise level $t$:
\begin{equation}
    \label{eq:diffusion_loss}\mathcal{L}_{\text{MSE}}=\|\mathbf{\epsilon} - \epsilon_\theta(\vx_0 +  \mathbf{\epsilon}^t, t))\|^2
\end{equation}
 To generate an image from the diffusion model, a sample $\vx_T$ at noise level $T$ is initialized from Gaussian noise $\mathcal{N}(0, 1)$. This sample $\vx_T$ is used for generation by iterative denoising:
\begin{equation}
    \vx^{t-1}=\vx^{t}-\gamma\epsilon_\theta(\vx^t,t) + \xi, \quad \xi \sim \mathcal{N} \bigl(0, \sigma^2_t I \bigl),
    \label{eq:unconditional_langevin}
\end{equation}
where $\gamma$ is the step size\footnote{An additional linear decay is further typically applied to the output $\vx^t$.}.
The final sample $\vx_0$ after denoising corresponds to a generated image.
The denoising function $\epsilon_\theta$ learns the score of an underlying EBM (unnormalized) probability distribution~\cite{vincent2011connection, song2019generative, liu2022compositional} and thus the above expression is equivalent to
\begin{equation}
    \vx^{t-1}=\vx^{t}-\gamma \nabla_{\vx} E_{\theta}(\vx^{t}) + \xi, \quad \xi \sim \mathcal{N} \bigl(0, \sigma^2_t I \bigl),
    \label{eq:ebm_langevin}
\end{equation}
where the denoising network $\epsilon_\theta(\vx^t,t)$ represents an unnormalized (EBM) density of data $p(\vx) \propto e^{-E_\theta(\vx)}$ by parameterizing $\nabla_{\vx} E_{\theta}(\vx)$ with the denoising function. This EBM interpretation of diffusion models enables the composition of different diffusion models together as discussed in \sect{sect:composable_diffusion} and further enables us to decompose images into multiple sets of different diffusion models.

\subsection{Composable Diffusion Models}
\label{sect:composable_diffusion}

Given two separate DDPM models $\epsilon_{\vc_1}$ and $\epsilon_{\vc_2}$ which parameterize two conditional EBM distributions~\cite{du2019implicit} $p(\vx|\vc_1)$ and $p(\vx|\vc_2)$ specifying the likelihood of images exhibiting concept $\vc_1$ and $\vc_2$, composable diffusion~\cite{liu2022compositional} proposes to generate images with both attributes by modifying the iterative denoising procedure using the hybrid denoising score $\epsilon_{\text{comb}}$:
\begin{equation}
    \vx^{t-1}= \vx^{t}-\gamma \bigl(\epsilon_{\text{comb}}(\vx^t,t)\bigl) + \xi, \quad \xi \sim \mathcal{N} (0, \sigma^2_t I).
    \label{eq:compose_langevin}
\end{equation}
The hybrid denoising function $\epsilon_{\text{comb}}$ corresponds to a composition of score functions:
\begin{equation}
    \epsilon_{\text{comb}}(\vx^t,t) = \epsilon_{\vc_1}(\vx^t,t) + \epsilon_{\vc_2}(\vx^t,t) - \epsilon_{\phi}(\vx^t,t),
\end{equation}
where $\epsilon_{\phi}$ corresponds to a DDPM representing the unconditional image distribution $p(\vx)$.  Sampling using this hybrid denoising function corresponds to sampling from the composite EBM distribution~\cite{liu2022compositional} \footnote{Assuming that $\vc_1$ and $\vc_2$ are independent.}:
\begin{equation}
    p(\vx|\vc_1, \vc_2) \propto \frac{p(\vx|\vc_1)p(\vx|\vc_2)}{p(\vx)}.
\end{equation}
This property of composable diffusion enables us to construct and sample from complex novel compositions of different concepts at test time. In this paper, we aim to infer a set of composable concepts from training images in an unsupervised manner.

\section{Method}

In this section, we introduce our unsupervised approach that discovers compositional concepts from a set of images using a pretrained diffusion model. We first formulate how we may decompose data points into unsupervised concepts with diffusion models. Next, we illustrate how we may infer these unsupervised concepts using learned latent representations (\ie, word embeddings) in a text-to-image generative model.

\subsection{Unsupervised Compositional Discovery}
\label{method:unsupevised_discovery}
Given a dataset of images $\{\vx_i\}$, we aim to discover a set of independent compositional concepts $\{\vc^1_i, \cdots, \vc^K_i\}$ for each image $\vx_i$ in an \textit{unsupervised manner}, each specifying a conditional EBM distribution $p(\vx_i|\vc_i^k)$, which represent different components of the image. In particular, the probability of each individual image $\vx_i$ can be decomposed as a product of $K$ independent concepts:
\begin{equation}
\resizebox{0.9\hsize}{!}{$
    p_{\text{decomp}}(\vx_i) = p(\vx_i|\vc_i^1, \ldots, \vc_i^K) \propto p(\vx_i) \prod_{k=1}^{K} \frac{p(\vx_i|\vc_i^{k})}{p(\vx_i)},$}
\end{equation}
where we represent each individual probability distribution $p(\vx|\vc_i^k)$ using a different denoising model $\epsilon_{k}(\vx^t, t)$.

Modeling this decomposed distribution $p_{\text{decomp}}(\vx_i)$ corresponds to sampling from the score function of a composite EBM~\cite{liu2022compositional}:
\begin{equation}
    \nabla_{\vx} E(\vx_i) + \sum_{k=1}^K (\nabla_{\vx} E(\vx_i|\vc_i^k) - \nabla_{\vx} E(\vx_i))
    \label{eqn:uncond_full_obj_energy}
\end{equation}
This then corresponds to constructing a new noise prediction model $\epsilon_{\text{unsup}}$:
\begin{equation}
    \epsilon_{\text{unsup}}(\vx_i^t,t) = \epsilon(\vx_i^t, t) + \sum_{k=1}^K \bigl(\epsilon(\vx_i^t, t|\vc_i^k) - \epsilon(\vx_i^t, t)\bigl),
    \label{eqn:uncond_full_obj}
    \vspace{-2mm}
\end{equation}
where $\epsilon(\vx_i^t, t)$ corresponds to the unconditional score prediction. To discover an independent set of compositional concepts for an image, we then wish to learn a denoising function such that for each image $\vx_i$ and noise $\epsilon^t$:
\begin{equation}
    \mathcal{L}_{\text{MSE}}=\|\mathbf{\epsilon}- \epsilon_{\text{unsup}}(\vx_i +  \mathbf{\epsilon}^t, t))\|^2.
    \label{eqn:denoise}
\end{equation}
To ensure that the set of decomposed concepts $\vc_i^k$ in each image is consistent across different images in our dataset, we parameterize each $\vc_i^k$ as the weighted sum $\vw_i^k$ of a library of $K$ concepts $\vc^k$ shared across all images. 
We optimize both a set of concepts $\vc^k$ as well as a set of per image/concept weight assignments $\vw_i^k$, where $\sum_k \vw_i^k = 1$ for each image $i$.
Our final modified score prediction corresponds to:
\begin{equation}
    \resizebox{0.87\hsize}{!}{$
    \epsilon_{\text{unsup}}(\vx_i^t,t) = \epsilon(\vx_i^t, t) + \sum_{k=1}^K \vw_i^k \bigl(\epsilon(\vx_i^t, t|\vc^k) - \epsilon(\vx_i^t, t)\bigl).
    $}
    \label{eqn:full_obj}
\end{equation}
This corresponds to representing each image with the product distribution
\begin{equation}
    p_{\text{decomp}}(\vx_i)  \propto p(\vx_i) \prod_{k=1}^{K} \left (\frac{p(\vx_i|\vc^{k})}{p(\vx_i)} \right )^{\vw_i^k}.
\end{equation}
When the vector of weights $\vw_i$ for each image is one-hot, images are ``clustered" into $K$ separate concepts $\vc^k$, where each image is represented by a single concept $\vc^k$ that represents its class identity (\ie dog or cat). In contrast, when the vector of weights $\vw_i$ is mixed across multiple different concepts $\vc^k$, each image can be decomposed into a set of factors representing multiple image attributes, such as scene lighting and objects. 

To discover concepts, we train $\epsilon_{\text{unsup}}$ on each image with the objective in \eqn{eqn:denoise} and jointly optimize per image weights $\vw_i$ and shared concepts $\vc^k$:
\begin{equation}
\begin{aligned}
    \vw_i &= \vw_i - \lambda \nabla_{\vw_i}\mathcal{L}_{\text{MSE}}, \\
    \vc^k &= \vc^k - \lambda \nabla_{\vc^k}\mathcal{L}_{\text{MSE}},
    \end{aligned}
\end{equation}
where $\lambda$ is the learning rate.
We provide the full pseudocode for discovering concepts in Algorithm~\ref{alg:training}.

\begin{figure}[t]
\vspace{-5pt}
\begin{minipage}{0.47\textwidth}
\begin{algorithm}[H]
    \centering
    \caption{Unsupervised Concept Discovery}
    \label{alg:training}
    \begin{algorithmic}[1]
        \STATE \textbf{Require} Diffusion model $\epsilon_{\theta}(\vx_i^t, t| c)$, training images $\{\vx_1, \ldots, \vx_N\}$, weights $ \{\vw_1, \ldots, \vw_N\}, \vw_i \in \R^{K}$, $K$ randomly initialized concept embeddings $\{\vc^1, \ldots, \vc^K\}$, learning rate $\lambda$. \\
        \FOR{$i = 0, \ldots, N$}
            \STATE Initialize a Gaussian noise $\epsilon \sim \mathcal{N}(0, 1)$ \\
            \STATE Initialize a noise $\epsilon^t = \alpha^t\epsilon$ at a random time step $t$ \\
            \STATE $\vx_i^t = \vx_i + \epsilon^t$ \hspace{2.1cm} \small{\color{gray}// add $ t$ levels of noise}\\
            \STATE $\epsilon_k \gets \epsilon_{\theta}(\vx_{i}^t, t | \vc^k)$ \hspace{0.4cm} \small{\color{gray}// compute $K$ conditional scores}  \\
            \STATE $\epsilon_\phi \gets \epsilon_{\theta}(\vx_i^t, t | \phi)$ \hspace{0.65cm} \small{\color{gray}// compute unconditional score}  \\
            \STATE $\epsilon_{\text{unsup}} \gets \epsilon_\phi + \sum_{k=1}^K \vw_i^k (\epsilon_k - \epsilon_\phi)$\hspace{1cm} \eqn{eqn:full_obj}
            \STATE $\mathcal{L}_{\text{MSE}}=\|\epsilon - \epsilon_{\text{unsup}}\|^2$ \hspace{1.2cm} 
            \small{\color{gray}// train score to denoise}
            \STATE \small{\color{gray}
            // update the weight $\vw_i$ and all $K$ concepts.}
            \STATE $\vw_i = \vw_i - \lambda \nabla_{\vw_i}\mathcal{L}_{\text{MSE}}$
            \STATE $\vc^k = \vc^k - \lambda \nabla_{\vc^k}\mathcal{L}_{\text{MSE}}$
        \ENDFOR \\
    \end{algorithmic}
\end{algorithm}
\end{minipage}
\end{figure}

\subsection{Parameterizing Concepts with Text-to-Image Generative Models}

In \cref{method:unsupevised_discovery}, we aim to construct a set of $K$ different conditional denoising networks $\epsilon(\vx_i^t, t|\vc^k)$ and an unconditional denoising network $\epsilon(\vx_i^t, t)$ which can jointly denoise images across our dataset. Directly discovering this score functions from scratch using a dataset of images is difficult as there may be substantial ambiguity on how images should be factored, with this difficulty compounded by the small dataset.

To more efficiently parameterize and discover these $K$ different score functions, we propose to parameterize each denoising prediction network $\epsilon(\vx_i^t, t|\vc^k)$ using a randomly initialized word embedding $\vc^k$ in a text-to-image diffusion model so that $\epsilon(\vx_i^t, t|\vc^k) = \epsilon_{\theta}(\vx_i^t, t | \vc^k)$. Parameterizing a denoising function in text embedding space is substantially lower dimensional than discovering the score function from scratch, enabling learning / concept discovery from limited sets of data. Furthermore, the semantic space of text eliminates a lot of ambiguity when discovering concepts.

Note, that while in our current implementation, we optimize each shared concept $\vc^k$ using a word embedding and a weight vector $\vw_i \in \R^{K}$ for each image $\vx_i$, we can parameterize these $K$ different denoising networks $\epsilon_\theta(\vx_i^t, t|\vc^k)$ in other ways. For instance, we can directly parameterize these $K$ score functions by optimizing all parameters of the text-to-image model per concept or by optimizing a small adapter in a similar fashion as~\cite{ruiz2022dreambooth} on the weights of each model.

\section{Experiments}

\begin{table*}[t]
    \centering
    \small
    \setlength{\tabcolsep}{2mm}
    \scalebox{1}{
    \begin{tabular}{l|cc|cc|cc|cc|cc}
        \toprule
        \bf \multirow{2}{*}{Models} 
        & \multicolumn{2}{c|}{\bf ImageNet $S_1$}
        & \multicolumn{2}{c|}{\bf ImageNet $S_2$} 
        & \multicolumn{2}{c|}{\bf ImageNet $S_3$} 
        & \multicolumn{2}{c|}{\bf ImageNet $S_4$} 
        & \multicolumn{2}{c}{\bf Average}\\
        \cmidrule{2-11}
        & \bf Acc $\uparrow$ & \bf KL $\downarrow$
        & \bf Acc $\uparrow$ & \bf KL $\downarrow$
        & \bf Acc $\uparrow$ & \bf KL $\downarrow$
        & \bf Acc $\uparrow$ & \bf KL $\downarrow$
        & \bf Acc $\uparrow$ & \bf KL $\downarrow$ \\
        \midrule
        \bf Textual Inversion~\cite{gal2022image} 
        & 4.06 & 0.5756
        & 7.19 & 0.1152
        & 36.88 & 0.1525 
        & 63.44 & 0.4958
        & 27.89 & 0.3348
        \\
        \bf Textual Inversion (KM)
        & 44.37 & 0.3799
        & 22.50 & \bf 0.0926
        & 37.81 & 0.2123
        & 78.75 & 0.3576
        & 45.86 & 0.2606
        \\ 
        \bf Textual Inversion (CKM)
        & 48.13 & \bf 0.0282 
        & 24.38 & 0.2367 
        & \bf 63.75 & 0.1569 
        & 69.38 & 0.2249 
        & 51.41 & 0.1617
        \\
        \bf Ours
        & \bf 56.88 & 0.1613 
        & \bf 26.56 & 0.2929
        & 56.56 & \bf 0.1323
        & \bf 82.81 & \bf 0.0285
        & \bf 55.70 & \bf 0.1538
        \\
        \midrule
        & \bf CLIP  $\uparrow$ & \bf KL $\downarrow$
        & \bf CLIP $\uparrow$ & \bf KL $\downarrow$
        & \bf CLIP  $\uparrow$ & \bf KL $\downarrow$
        & \bf CLIP $\uparrow$ & \bf KL $\downarrow$
        & \bf CLIP $\uparrow$ & \bf KL $\downarrow$ \\
        \midrule
        \bf Textual Inversion~\cite{gal2022image} 
        & 13.13 & 0.4195
        & 3.44 & 0.1182
        & 9.38 & \bf 0.0554
        & 33.75 & 0.7193
        & 14.93 & 0.3281
        \\
        \bf Textual Inversion (KM)
        & 29.06 & 0.1299
        & 9.38 & \bf 0.0803 
        & 12.19 & 0.3325 
        & 44.37 & 0.3799
        & 23.75 & 0.2307
        \\
        \bf Textual Inversion (CKM)
        & 40.31 & \bf 0.0020 
        & 10.00 & 0.3384 
        & 14.37 & 0.1490 
        & \bf 69.69 & 0.4232 
        & 33.59 & 0.2281
        \\
        \bf Ours
        & \bf 42.19 & 0.2091
        & \bf 30.00 & 0.1518
        & \bf 30.63 & 0.1513
        & 51.56 & \bf 0.0224
        & \bf 38.60 & \bf 0.1337
        \\
        \bottomrule
    \end{tabular}}
    \vspace{-5pt}
    \captionof{table}{\small \textbf{Quantitative Evaluation of Discovered Concepts.} We assess the accuracy of decomposed concepts in capturing each ImageNet class in the data using pre-trained ResNet-50 and CLIP classifiers. We also report the KL divergence of discovered classes.}
    
    \label{table:classification}
\end{table*}

In this section, we compare our approach with baseline methods in concept discovery on three different tasks, including object discovery, indoor scene discovery, and artistic concept discovery. We further show the results of compositional image generation and representation learning using the discovered concepts. We also provide visualizations and analysis on sensitivity and robustness in Appendix~\sect{supp:additional_results}.

\subsection{Datasets}
\label{exp:datasets}
\paragraph{ImageNet~\cite{deng2009imagenet}.} For the experiment, we select $4$ sets of class combinations, denoted as ImageNet $S_1$, $S_2$, $S_3$ and $S_4$.  Each set consists of $5$ classes from ImageNet, with $5$ randomly chosen images per class. During testing, we generate $64$ images per concept for evaluation. 

\paragraph{ADE20K~\cite{zhou2017scene}.} 
In this experiment, our goal is to discover concepts from \textit{kitchen} images in the ADE20K dataset. We randomly select $25$ images as the training data. 

\paragraph{Artistic Paintings.} 
To further demonstrate the ability of our method to discover a wide range of visual concepts, we collected a dataset of artistic paintings from the internet. The dataset includes $5$ paintings by Van Gogh, $7$ painting samples by Claude Monet, and $5$ painting images by Pablo Picasso.

\subsection{Evaluation Metrics}
\label{exp:eval_metrics}

\paragraph{Classification Accuracy.} 
To evaluate the effectiveness of each method on decomposing unlabeled ImageNet images into a set of meaningful classes, we utilize a pre-trained classification model to compute image classification accuracy.
For each class present in the training data,  we feed the generated images into the pre-trained ResNet-50 and extract the largest value from the logit values of target classes (\ie, the $5$ class targets in our setting).
To evaluate the accuracy of our model predictions, we establish a threshold of $10$ for the logit value. Predictions with logit values above this threshold are counted as correct, while those below are deemed incorrect. We find that this threshold represents a high prediction confidence from ResNet-50. Finally, we report the per-set accuracy and average accuracy across all the sets as our final results.

\paragraph{CLIP Accuracy.} To further evaluate the accuracy of decomposed concepts in ImageNet, we compute a CLIP accuracy using the pairwise CLIP similarity score between generated images from each concept with the set of classes present in the training data, using pre-trained CLIP encoders~\cite{radford2021learning}. The highest-scoring caption is then selected as the model prediction. We count a prediction as correct if the CLIP similarity score is greater than a certain threshold (\ie, $0.3$ in our experiments).

\paragraph{KL Divergence.} 
We use Kullback-Leibler divergence to further evaluate the effectiveness of capturing diverse image concepts on ImageNet. This dataset primarily consists of object concepts, making it an ideal choice for measuring and illustrating the differences we are interested in. Ideally, each decomposed concept should reflect a separate class in the data, resulting in an equal number of generated images per class. We assign an image to a class based on the class label that receives the highest logit value. We then compute the KL divergence between the distribution of classes inferred in this manner with a uniform distribution of classes in the training data (ground truth). In our experiments, we calculate KL divergence by computing logits with regards to both pre-trained ResNet-50 and CLIP encoders.

\paragraph{Representation Accuracy.}
Our proposed method aims to discover a set of concept representations, which can be further used for downstream tasks, such as classification. Thus, we also evaluate the quality of representations using clustering for classification. First, we utilize K-means clustering, where we assign a cluster to an ImageNet class based on the most frequent class of training images in that cluster. In test time, we count each test example as correct if it is assigned to the cluster with the same image label.

\noindent We use the pretrained Stable Diffusion v2.1 model in our experiments. For image generation, we utilize classifier-free guidance~\cite{ho2022classifier} to sample $64$ images for each ImageNet class with $50$ steps using the DDIM sampler~\cite{song2020denoising}.%

\begin{figure}[t]
\begin{center}
\includegraphics[width=0.47\textwidth]{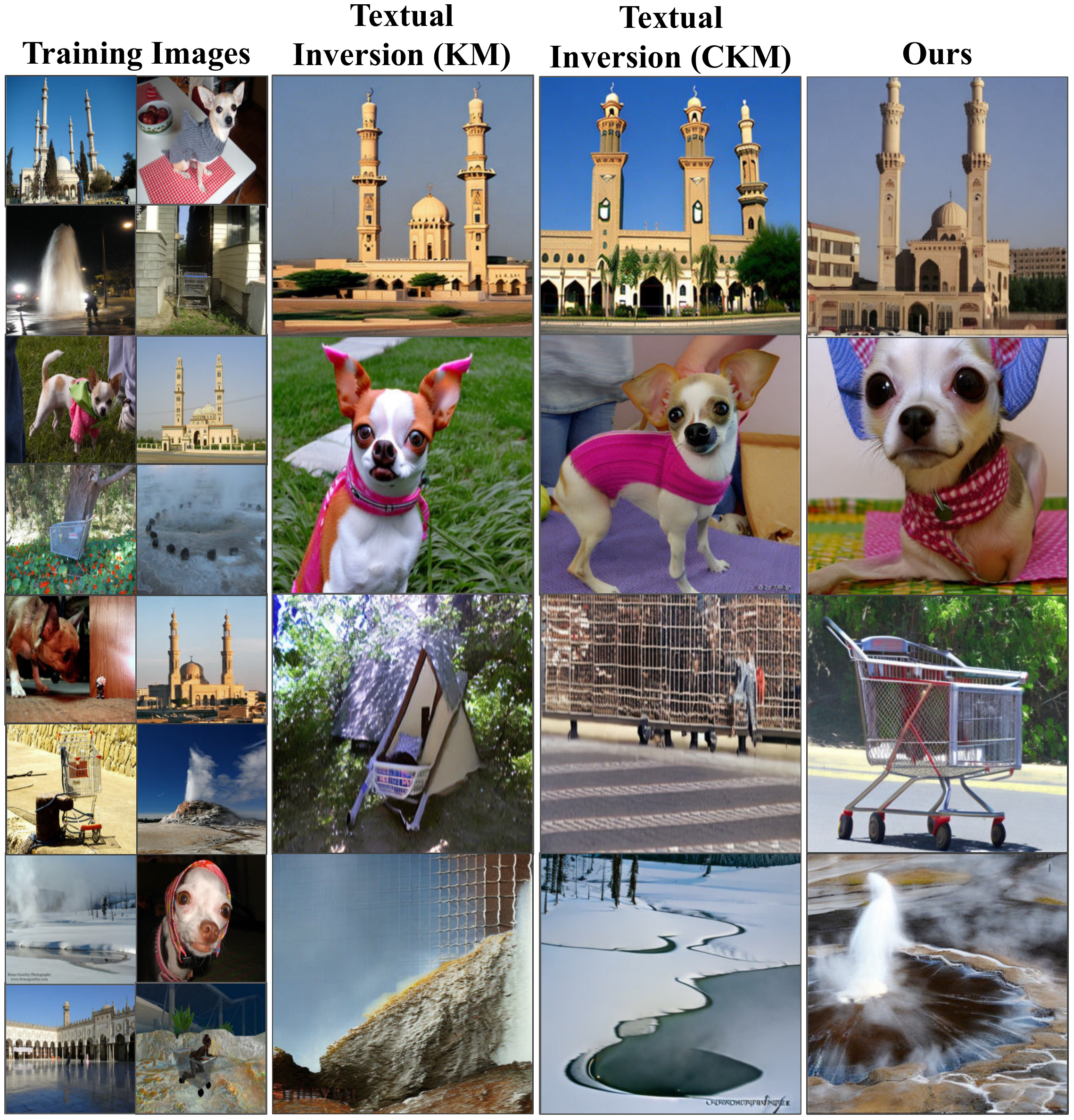}
\end{center}
\vspace{-15pt}
\caption{\small \textbf{Object Decomposition.} Our proposed method can discover different object categories from a set of unlabeled images.
}
\label{fig:imagenet_decomposition}
\end{figure}

\subsection{Baselines}
\label{exp:baselines}
\paragraph{COMET}~\cite{du2021comet} is the first work that utilizes a composite of EBMs to decompose images into a set of concepts in an unsupervised manner, but it scales poorly to more complex in-the-wild images. In contrast, our approach enables automatic concept discovery across in-the-wild images.

\paragraph{Textual Inversion}~\cite{gal2022image} is one of the first works to utilize the text-to-image diffusion model to learn a mapping from a set of similar images to a single-word representation. Unlike our unsupervised method, textual inversion optimizes a single representation using a set of similar images, thus assuming a correspondence between training images and the target word representation. In contrast, our method enables unsupervised learning of multiple concepts simultaneously in one single training run. To provide a fair comparison, we developed a baseline using textual inversion to map all images into an unconditional word representation. Each ImageNet set has $5$ distinct classes, so the single unconditional representation may ideally learn a uniform distribution of all $5$ image concepts. For evaluation, we sample $320$ images for classification accuracy. In contrast, we sampled $64$ samples for each of the $5$ concepts, thus a total of $320$ images, in our method for ImageNet dataset.

\paragraph{Textual Inversion + K-means} is a modified version of textual inversion~\cite{gal2022image}. Since images are unlabeled, we utilize K-means clustering~\cite{lloyd1982least} to obtain pseudo-labels. In our experiments, we use two variants of K-means clustering: K-means (KM) in pixel space and CLIP-based K-means (CKM). We first utilize K-means clustering to obtain pseudo-labels for the given training images, and then train textual inversion on such image-label pairs. 

\subsection{Unsupervised Concept Discovery}

Our method can decompose images from different domains into concepts, including objects, components in indoor scenes and artistic styles, without using any labels.

\subsubsection{Object Discovery}

We show that our proposed method can automatically discover object concepts from a set of unlabeled images.

\begin{figure}[t]
\begin{center}
\includegraphics[width=0.47\textwidth]{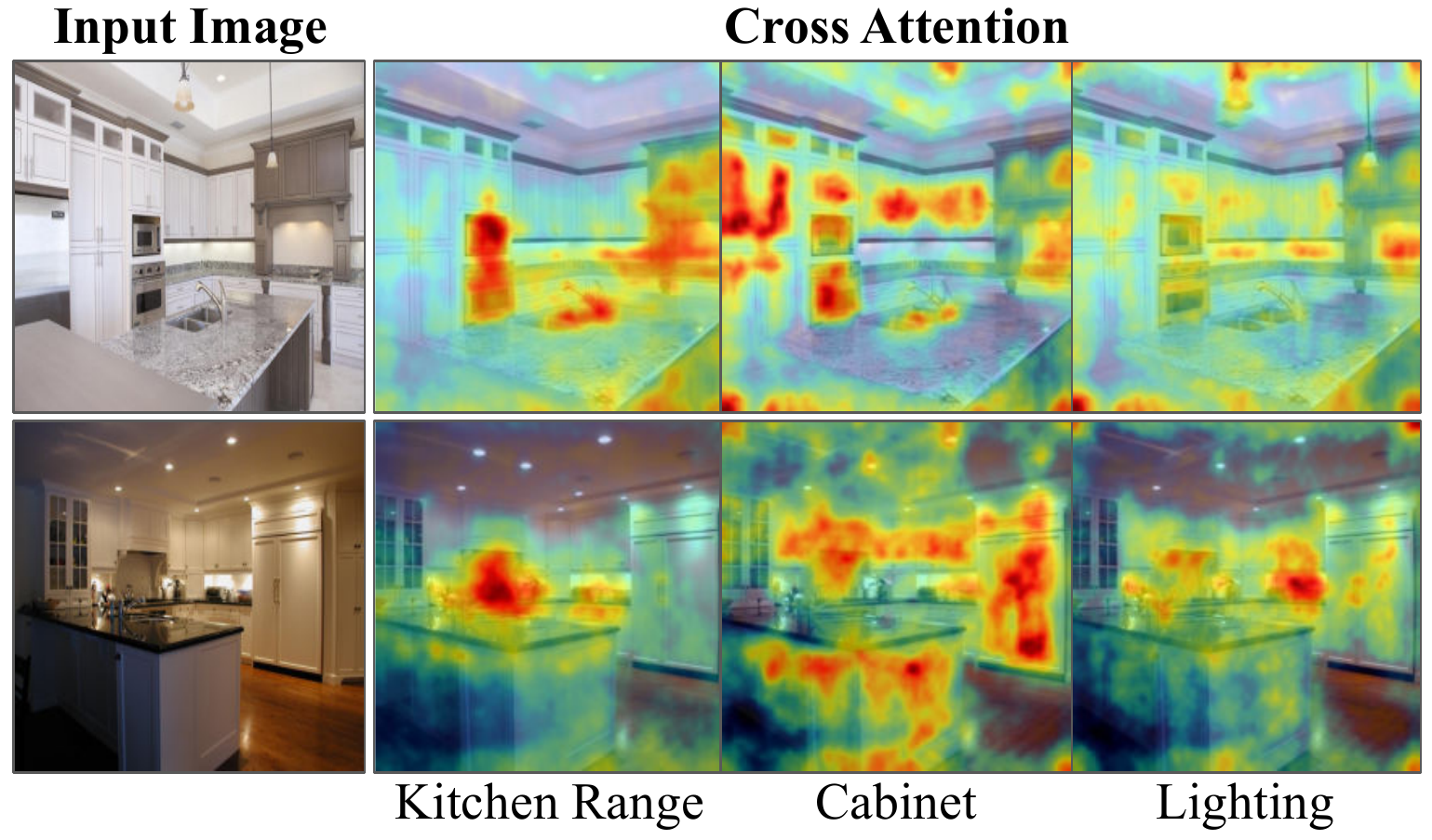}
\end{center}
\vspace{-15pt}
\caption{\small \textbf{Cross Attention Concept Visualization.} We visualize the attention maps of three discovered concepts from unlabelled images. The concepts focus on different portions of the dataset.
}
\label{fig:ade20k_training_vis}
\end{figure}

\begin{figure}[t]
\begin{center}
\includegraphics[width=0.47\textwidth]{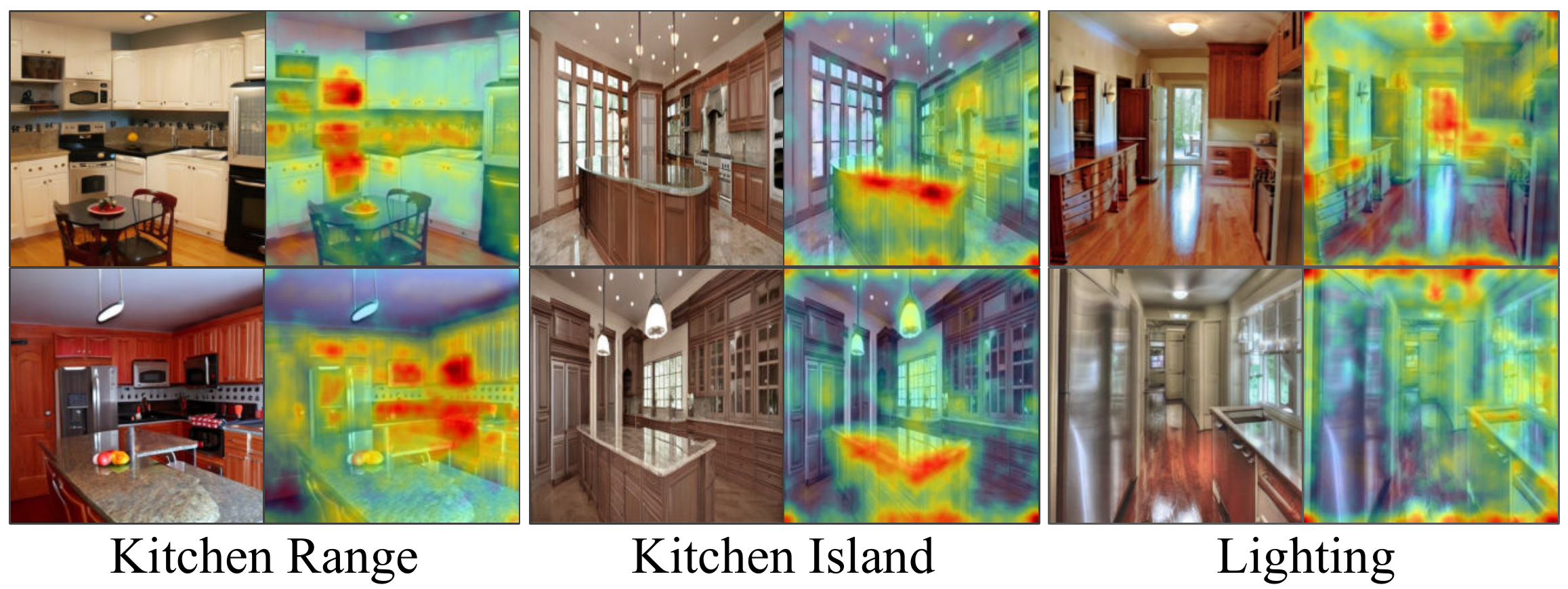}
\end{center}
\vspace{-15pt}
\caption{\small \textbf{Kitchen Scene Decomposition.} We show generated images (odd columns) along with the attention maps of the corresponding concept (even columns) with respect to the image.
}
\label{fig:ade20k_decomposition}
\end{figure}

\begin{figure*}[t]
\small
\centering
\includegraphics[width=1\linewidth]{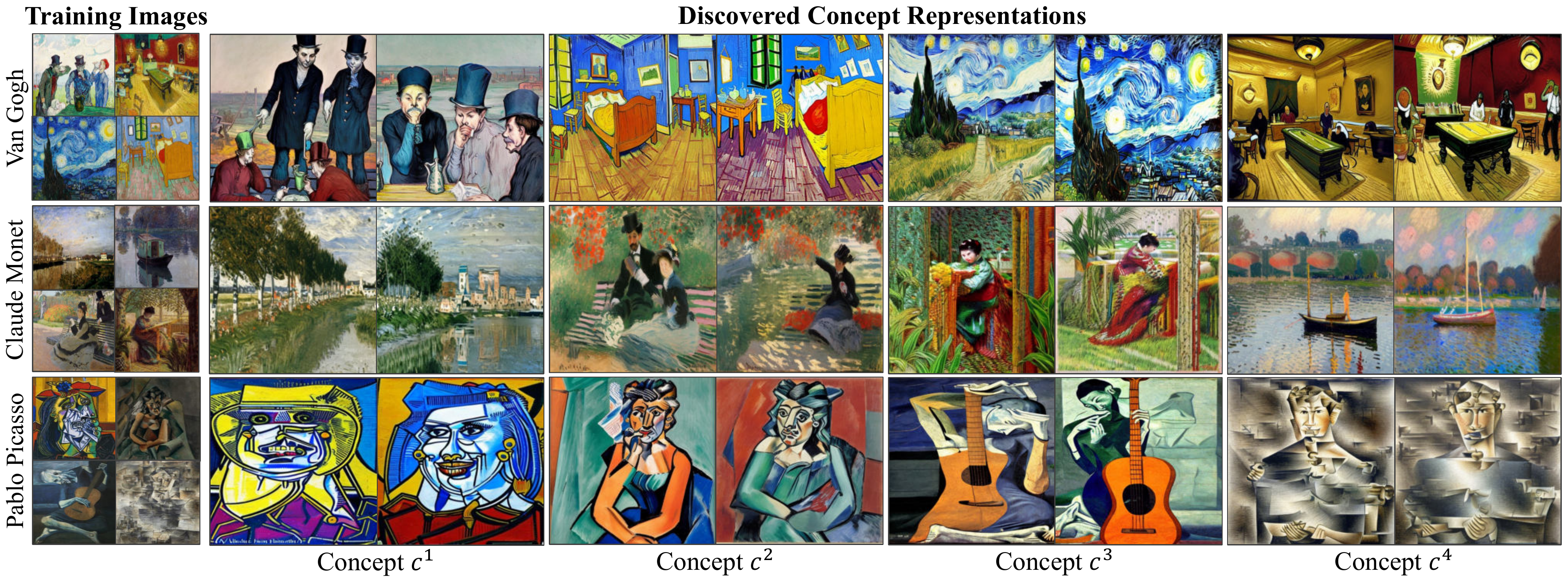}
\vspace{-15pt}
\caption{\small
\textbf{Unsupervised Concept Decomposition on Arts.} Our method allows unsupervised concept decomposition from just a few paintings (\ie, $5-7$ per artist), with each concept $c^i$ representing a distinct concept. For instance, in the first row, $c^1$ represents ``drinkers'', while in the third row, $c^3$ represents ``guitarist''.
}
\label{fig:art_decomposition}
\end{figure*}

\paragraph{Qualitative results.} We first demonstrate that our method can faithfully decompose ImageNet images into a set of object concepts. We qualitatively compare our method with multiple variants of textual inversion~\cite{gal2022image} in \fig{fig:imagenet_decomposition}. Although all three methods achieve similar performance on ``mosque'' and ``Chihuahua'' in the top two rows, both clustering-based textual inversion methods (KM and CKM) fail to capture visual concepts of ``shopping cart'' and ``geyser'' in the bottom two rows. In contrast, our method can capture all four concepts faithfully.
We further evaluate COMET~\cite{du2021comet} on this setting, but find that it obtains low performance in our setting as it fails to generate photorealistic images. See \cref{supp:additional_results} for additional qualitative results of COMET and other approaches.

\paragraph{Quantitative results.} In Table~\ref{table:classification}, we compare our method with baselines quantitatively using image classification accuracy.
Our proposed method achieves a higher or comparable classification performance across different sets of ImageNet combinations using pre-trained ResNet-50~\cite{he2016deep} and CLIP~\cite{radford2021learning}. Furthermore, we evaluate the diversity of discovered concepts using discrete KL divergence between prediction distribution and target distribution. Intuitively, a lower KL divergence value indicates that the probability distribution of the generated images is closer to the uniform distribution, thus implying greater diversity in the generated images.
Compared to the baselines, our method achieves a consistently low KL divergence across different ImageNet sets. In contrast, the baselines exhibit a wider range of KL scores, suggesting that our proposed approach is more stable in terms of learning diverse concepts across these sets. As shown in the rightmost column of Table~\ref{table:classification}, the averaged results across all four sets further show that our method achieves the best performance on both accuracy and KL divergence, indicating its ability to learn diverse concepts.

\subsubsection{Indoor Scene Discovery}
To further verify the effectiveness of our approach, we demonstrate our method can decompose kitchen scenes into multiple sets of factors.

\paragraph{Qualitative results.} We evaluate our method on concept discovery for indoor scenes, specifically kitchen scenes from ADE20K~\cite{zhou2017scene}. Since our method discovers concepts in an unsupervised manner, there is no label for the learned concepts. Thus, we utilize Diffusion Attentive Attribution Maps (DDAM) ~\cite{tang2022daam} to visualize the relation between learned concepts and image contents. Specifically, DAAM utilizes word-pixel scores from cross attention layers to generate heap maps for visualization. As shown in \fig{fig:ade20k_training_vis}, we obtain DDAM associated with each concept by running DDIM inversion~\cite{song2020denoising} on the training image. Our method can decompose the kitchen scenes into different components such as kitchen range (\ie, stove and microwave), cabinets, and lighting effects. Furthermore, we visualize images that are generated conditioned on each individual inferred concept in \fig{fig:ade20k_decomposition}. Both figures show that our model can decompose challenging kitchen scenes into a set of meaningful factors.

\subsubsection{Artistic Concept Discovery}

Art has been a long-standing topic being studied in the computer vision and computer graphics community. Here we provide qualitative evaluations in the artistic domain to further demonstrate the versatility of our method.

\paragraph{Qualitative results.} As shown in \fig{fig:art_decomposition}, we demonstrate concept decomposition on artistic paintings from different artists, including Van Gogh, Claude Monet, and Pablo Picasso. In the second row, we show that our model can decompose training images into different concepts, including ``trees on the side", ``a lady sitting on a bench'', ``an embroidering lady'' and ``a boat'', with similar artistic styles to the original images. In the next section, we will discuss that the discovered concepts can be further composed together to generate images.

\subsection{Composing Discovered Concepts}

\label{exp:concept_composition}

After a set of factors is discovered from a collection of images, our method can further compose these concepts for compositional image generation using compositional operators from previous works~\cite{du2020compositional,liu2022compositional}. As our approach is unsupervised and there is no label for the discovered concepts, we add names for the discovered concepts manually for easy understanding. As shown in the previous section, our method can decompose images into meaningful concepts, such as objects in indoor scenes and artistic styles. These concepts can be further composed with other concepts to generate images with specific styles.

\begin{figure}[t]
\begin{center}
\includegraphics[width=0.47\textwidth]{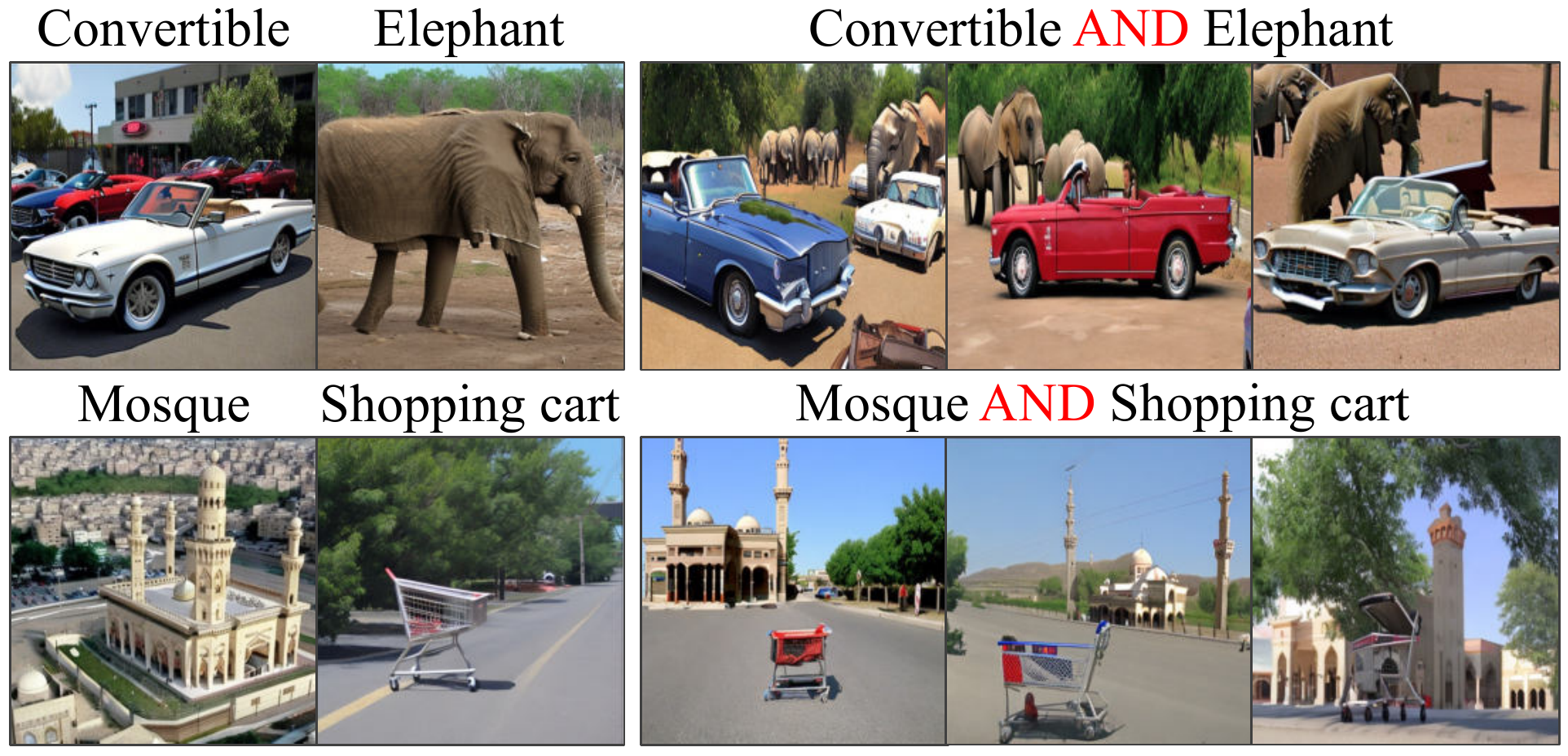}
\end{center}
\vspace{-20pt}
\caption{\small \textbf{Object Composition.} Our proposed method can generate images that showcase a composition of object concepts.
}
\label{fig:imagenet_composition}
\end{figure}

\paragraph{Object Composition.}
We are able to use the conjunction operator (\eg, AND) from Composable-Diffusion~\cite{liu2022compositional} to generate images that contain combinations of concepts that are unseen during training. 
As shown in Figure \ref{fig:imagenet_composition}, we show examples of generated images with the combination of a convertible car and elephants, as well as a composition of a shopping cart and a mosque.

\begin{figure}[t]
\begin{center}
\includegraphics[width=0.47\textwidth]{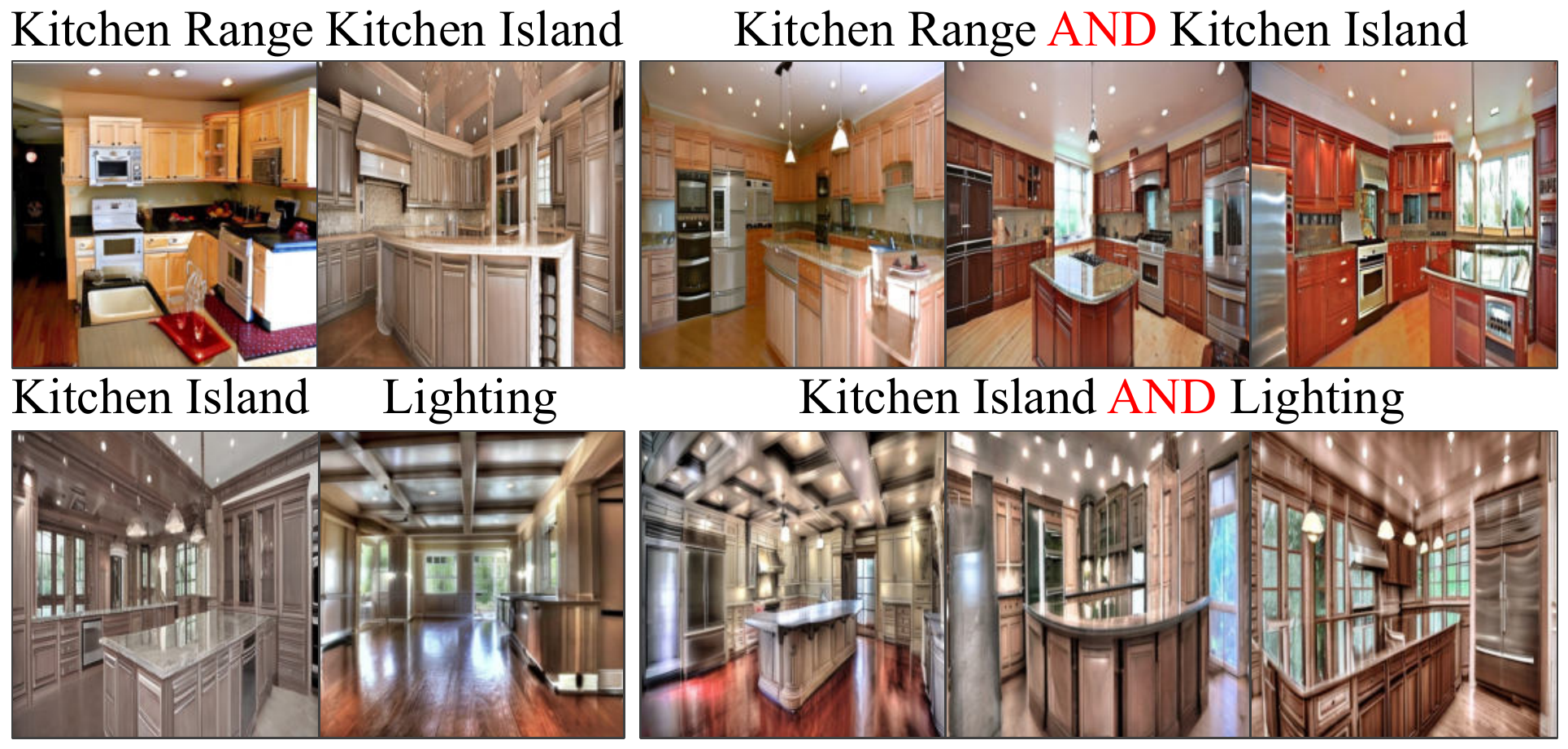}
\end{center}
\vspace{-15pt}
\caption{\small \textbf{Kitchen Concept Composition.} Our method demonstrates the ability to compose different components, such as kitchen ranges and lighting effects.
}
\label{fig:ade20_composition}
\end{figure}

\paragraph{Scene Composition.}
We further demonstrate the proposed method can enable scene composition using discovered concepts other than objects in the kitchen setting. As shown in Figure \ref{fig:ade20_composition}, our approach can discover concepts such as lighting and kitchen islands, and generate scenes with the specified objects and lighting effects.

\begin{figure}[t]
\begin{center}
\includegraphics[width=0.47\textwidth]{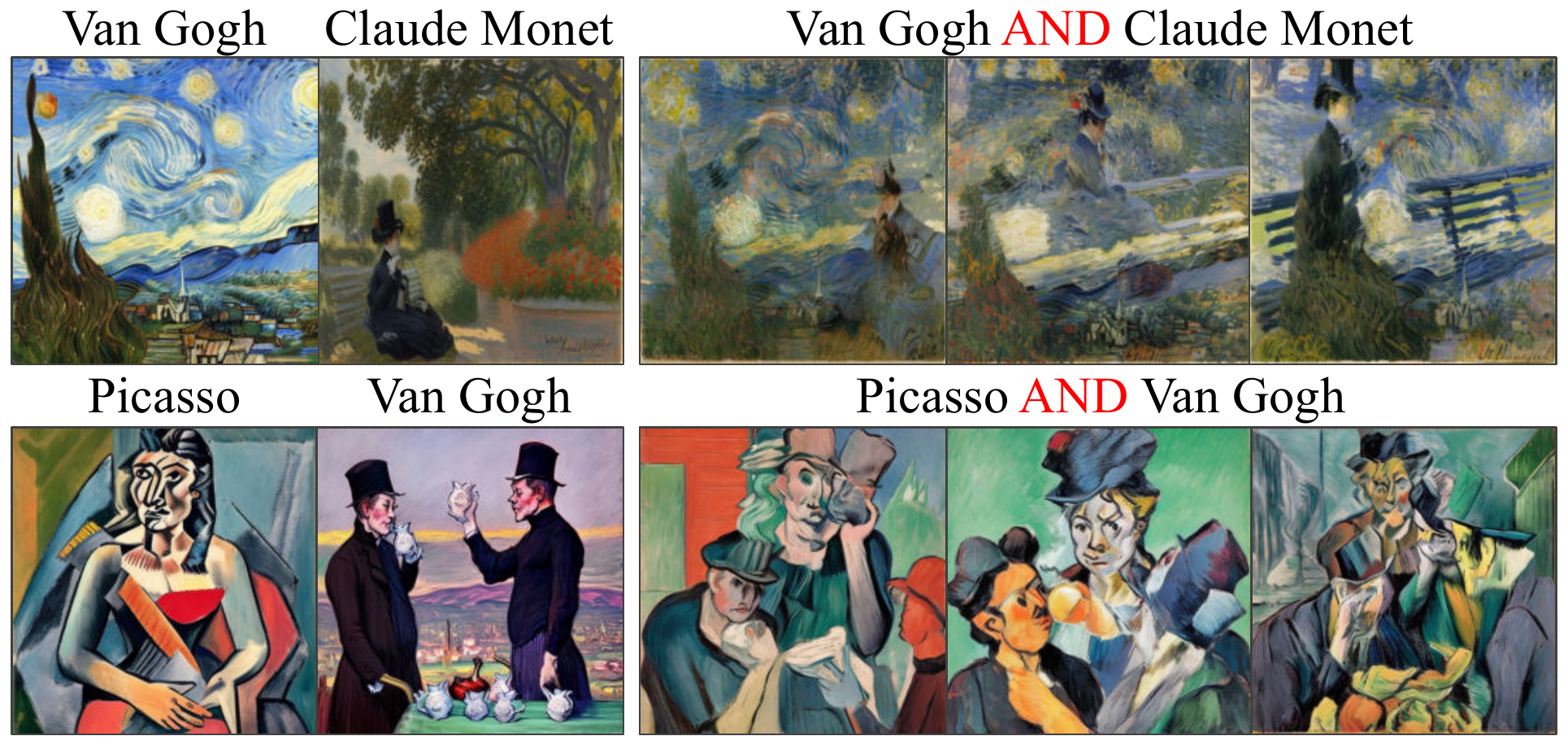}
\end{center}
\vspace{-15pt}
\caption{\small \textbf{Style Composition.} Our proposed method can compose artistic concepts learned from paintings, with each style named after the respective artist for better understanding.
}
\label{fig:art_style_composition}
\end{figure}

\paragraph{Style Composition.}
We can also combine artistic concepts discovered from paintings to generate images. As shown in Figure \ref{fig:art_style_composition}, we compose two types of discovered artistic styles to generate images using the conjunction operator. For example, images in the first row combine Van Gogh's starry night with Claude Monet's Camille Monet. Images in the second row combine Van Gogh's drinkers with the Cubism style of Picasso.

\begin{figure*}[t]
\small
\centering
\includegraphics[width=1\linewidth]{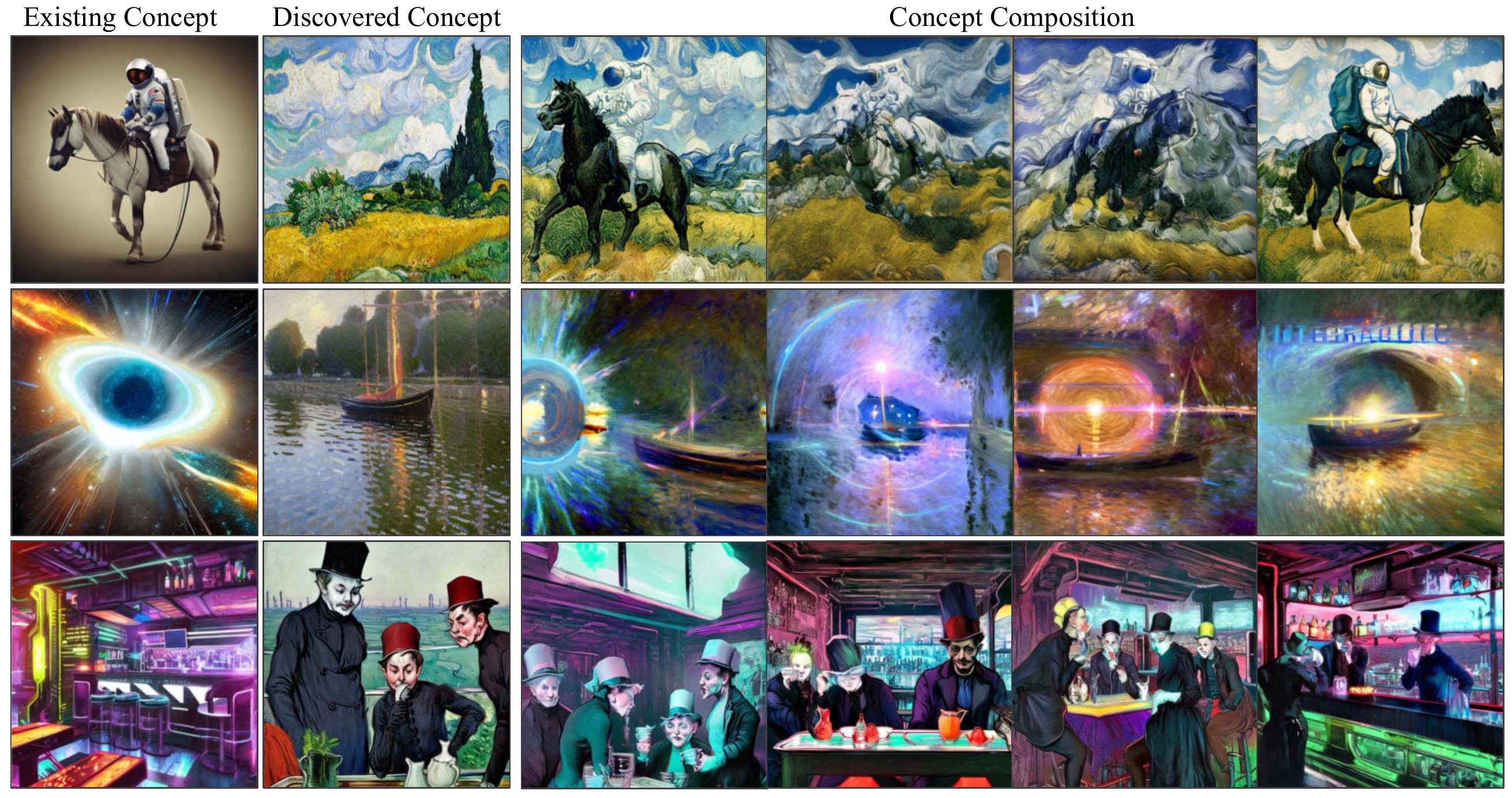}
\vspace{-20pt}
\caption{\small
\textbf{External Composition.} We demonstrate the ability to seamlessly integrate discovered concepts with existing concepts (text) to generate images with specified factors. For clarity, we omit the text descriptions. For instance, the \nth{1}, \nth{2}, \nth{3} images in the \nth{1} column are created using the phrase ``an astronaut riding a horse''. ``intergalactic wormhole'' and ``cyberpunk bar'', respectively.
}
\label{fig:art_composition}
\end{figure*}

\paragraph{External Composition.} Finally, our method can combine discovered concepts with external or existing knowledge from pre-trained generative models to generate images with new combinations. 
As shown in Figure \ref{fig:art_composition}, we combine textual descriptions with discovered concepts to create images that depict ``an astronaut riding a horse`` AND wheat field in the first row, ``intergalactic wormhole`` AND a boat in the second row, and ``cyberpunk bar'' AND drinkers in the last row, where the former is the text input, the latter is our discovered concepts and ``AND'' is the conjunction operator.

\begin{table*}[t]
    \centering
    \small
    \setlength{\tabcolsep}{3mm}
    \scalebox{0.9}{
    \begin{tabular}{l|c|c|c|c|c}
        \toprule
        \bf Methods
        & \bf ImageNet $S_1$ $\uparrow$
        & \bf ImageNet $S_2$ $\uparrow$
        & \bf ImageNet $S_3$ $\uparrow$
        & \bf ImageNet $S_4$ $\uparrow$
        & \bf Average $\uparrow$ \\ 
        \midrule
        \bf K-means
        & 37.00
        & 34.00
        & 37.00
        & 21.00
        & 32.25
        \\
        \bf Textual Inversion
        & 24.00
        & 24.00
        & 25.00
        & 24.00
        & 24.25
        \\
        \bf Ours
        & 58.00
        & 51.00
        & 59.00
        & 83.00
        & \bf 62.75
        \\
        \midrule
        \bf K-means (CLIP)
        & 65.00
        & 77.00
        & 87.00
        & 65.00
        & \bf 73.50
        \\
        \bottomrule
    \end{tabular}}
    \vspace{-5pt}
    \captionof{table}{\small \textbf{Generative Representation Evaluation.} Generative representations of images learned by our method can accurately classify images. K-means (CLIP) is a  supervised method and achieves the best average result.}
    \label{tab:rep_learning}
\end{table*}

\subsection{Representation Learning} 
By decomposing images into a weighted combination of
 compositional concepts, our approach discovers a representation for each training image. We can further obtain a representation for a test image $\vx_j$ by optimizing 
\eqn{eqn:denoise} and obtaining a weight vector $\vw_j \in \R^K$ for the image, while freezing the discovered concept representations $\vc^k$. We assess how this representation can be used for downstream tasks such as classification.

\paragraph{Quantitative Results.} We evaluate the effectiveness of the representation learned by our model for image classification. First, we use the optimized weight $\vw_i$ for each training image $\vx_i$ to fit a logistic regression model that predicts the ImageNet class based on the weight representation for each training image. We then evaluate the accuracy of the model on test images $\vx_j$ using the optimized weights $\vw_j$.

In this experiment, we compare our method with two variants of K-means clustering methods, one in pixel space and another in CLIP space. We also fit a logistic regression model on representations per image found using textual inversion~\cite{gal2022image}. We evaluate the accuracy of these methods for predicting ImageNet class on a test set of $100$ images. As shown in Table~\ref{tab:rep_learning}, our method achieves the best performance of $62.75\%$ mean accuracy compared to all the other unsupervised methods.
The method of CLIP-based K-means clustering is better than our method because the CLIP representation is directly trained in a supervised way on millions of image-text pairs.

\section{Conclusion}
We presented an approach to decompose datasets of images into a set of compositional generative concepts. 
Our approach is effective across a variety of datasets, including artistic paintings, indoor scenes, and ImageNet images. Additionally, we illustrated how discovered generative concepts can be combined with both each other and external concepts to generate novel images. Finally, we illustrated how discovered generative concepts can serve as a representation of an image which can be used for downstream tasks such as image classification. We hope our work opens a new direction of research on how generative models may not only be used to generate images but also as way to understand and represent images.

\newpage
{\small
\bibliographystyle{ieee_fullname}
\bibliography{egbib}
}

\clearpage
\appendix
\appendix

\noindent\textbf{\Large{Appendix}}
\vspace{10pt}

In this appendix, we first show additional results of both decomposition and composition in \cref{supp:additional_results}. We then provide details of datasets used in our experiments in \cref{supp:datasets}. Finally, we demonstrate details of baselines in \cref{supp:baselines} and our method in \cref{supp:our_approach}, respectively.

\section{Additional Results}
\label{supp:additional_results}

In this section, we first provide analyses of the performance of our method on the sensitivity of the number of concepts $K$, the variance of our method on inferred concepts, and the diversity of generated images in \cref{supp:analysis}. 
We then show additional results of decomposed concepts for objects, indoor scenes, artistic paintings, and hybrid dataset that consists of different modalities in \cref{supp:concept_discovery}. Finally, we provide additional results for object composition, indoor scene composition, art composition, and external composition in \cref{supp:composing_concepts}. Note that we utilize the conjunction operator (\eg, AND) from composable diffusion~\cite{liu2022compositional} for compositional generation.

\subsection{Analysis}
\label{supp:analysis}
\begin{figure}[h]
\begin{center}
\includegraphics[width=0.48\textwidth]{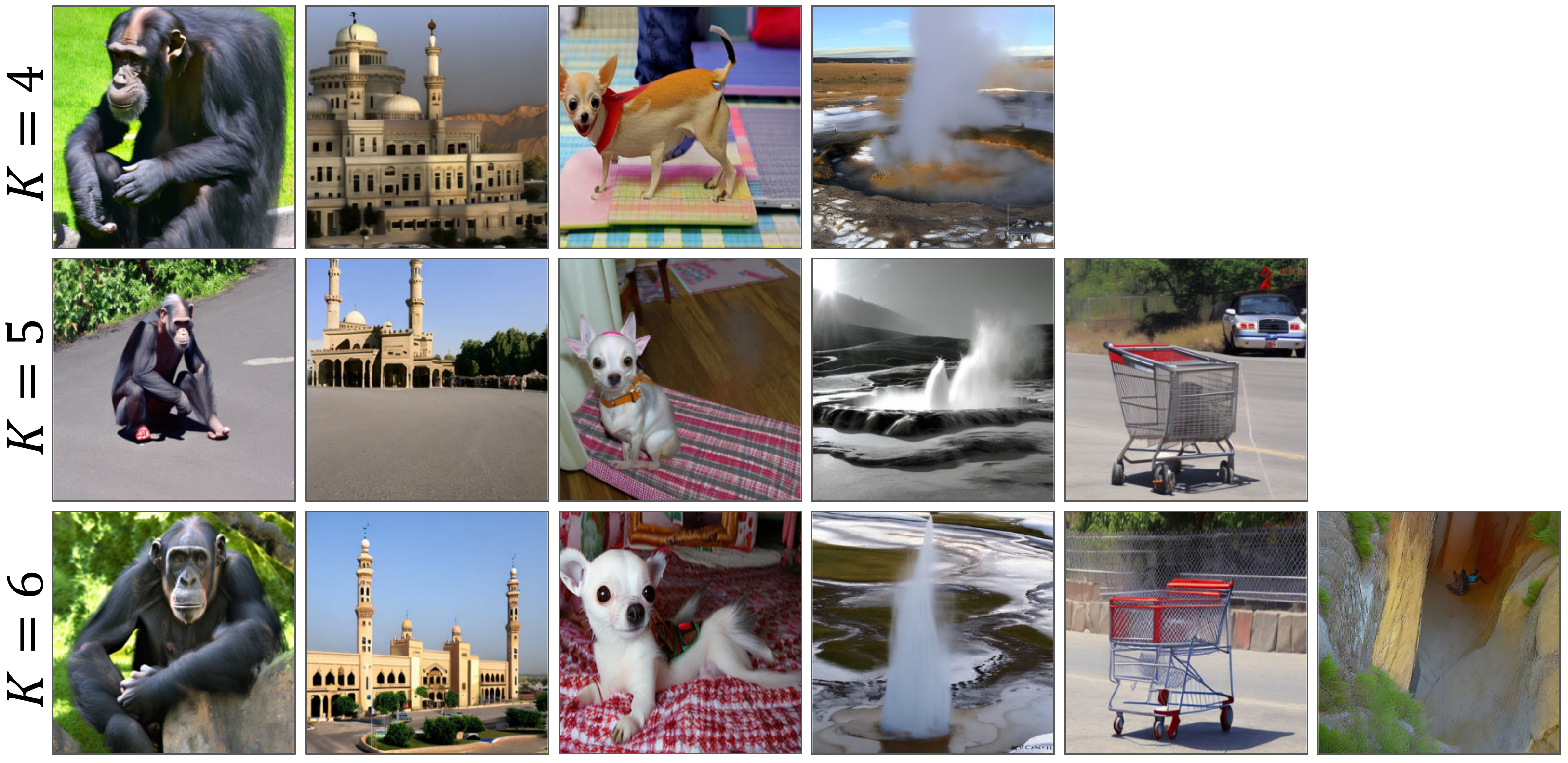}
\end{center}
\vspace{-15pt}
\caption{\small \textbf{Sensitivity of the number of concepts $\bm{K}$ on ImageNet Subset $\bm{S_1}$.}}
\label{fig:sensitivity_rebuttal}
\end{figure}

\paragraph{Sensitivity of the number of concepts $K$.}
In Figure~\ref{fig:sensitivity_rebuttal}, we run our method with varying values of $K$ (4, 5, and 6) on ImageNet $S_1$ (images with five categories of objects). When $K=5$, our method can correctly find all five concepts. When $K<5$, our method selects the top K obvious concepts. When $K>5$, our method tries to discover some new concepts from the training data.

\begin{figure}[t]
\begin{center}
\includegraphics[width=0.47\textwidth]{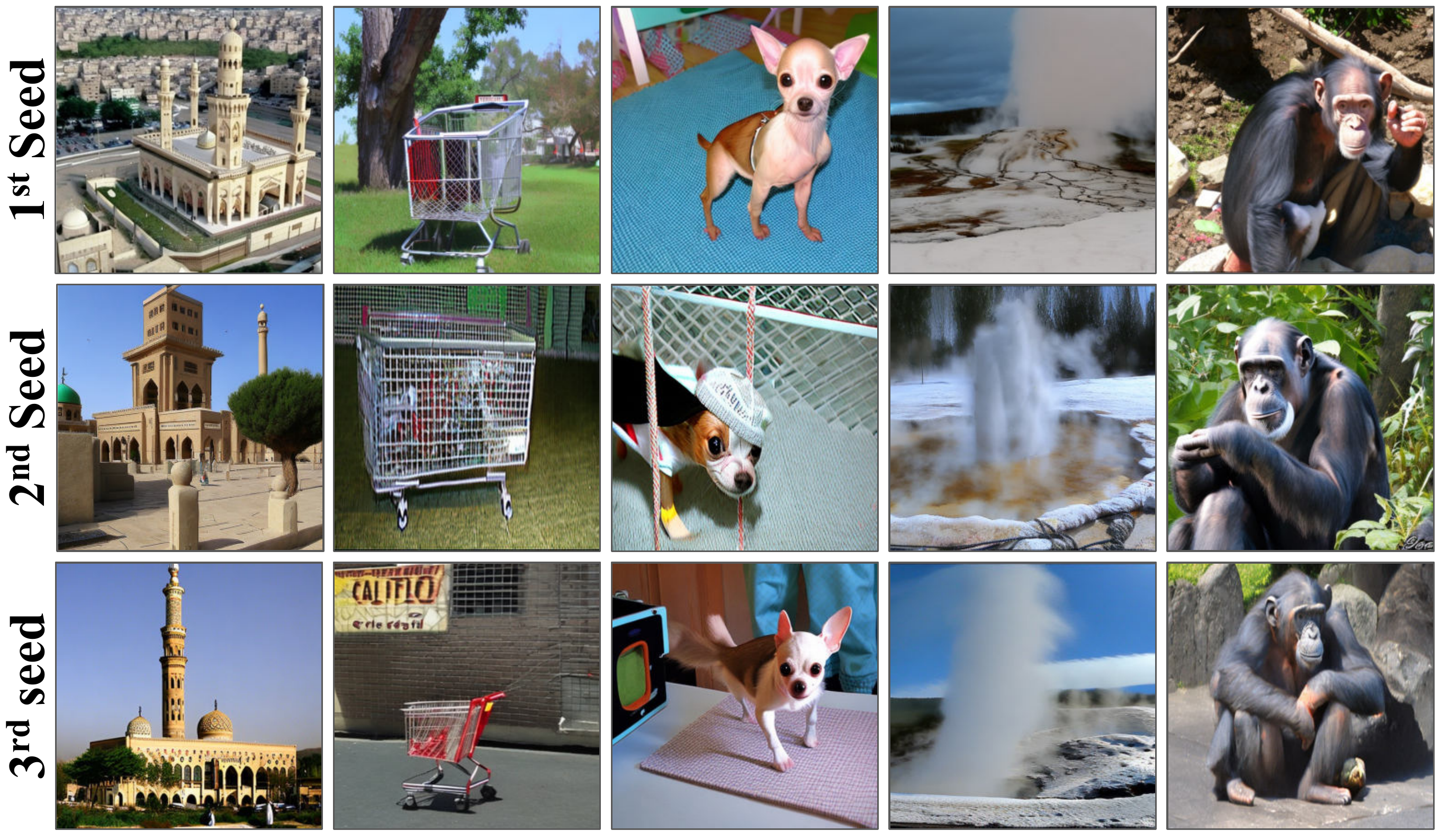}
\end{center}
\vspace{-15pt}
\caption{\small \textbf{Qualitative results on ImageNet $\bm{S_1}$ across $3$ random seeds.}
}
\label{fig:imagenet_decomposition_rebuttal}
\end{figure}

\paragraph{Variance of inferred concepts.}
We provide both qualitative and quantitive results on ImageNet $S_1$ across different seeds to assess the variance of inferred concepts. In Figure~\ref{fig:imagenet_decomposition_rebuttal}, we show that our method can reliably discover all object categories in $S_1$ across different seeds. In Table~\ref{table:classification_rebuttal}, we compare our method with the best baseline, \ie, Textual Inversion (CKM) on ImageNet $S_1$. The result shows that our method can capture concepts consistently across multiple runs, as evidenced by higher accuracy, lower KL Divergence and smaller standard deviation values. The result is also consistent with the qualitative results in Figure~\ref{fig:imagenet_decomposition_rebuttal}.

\paragraph{Diversity of generated images.} In Table~\ref{table:clip_rebuttal}, we measure diversity in both training data and generated images by computing pairwise dot product similarity using CLIP, since CLIP is trained to differentiate similar and dissimilar data, thus allowing us to measure diversity or dissimilarity. Our result shows that greater similarity in training images leads to less diversity in the generated images.

\begin{table}[t]
    \centering
    \small
    \setlength{\tabcolsep}{1mm}
    \scalebox{0.70}{
    \begin{tabular}{c|cc|cc}
        \toprule
        \multirow{2}{*}{\bf Models}  & \multicolumn{2}{c|}{\bf ResNet-50} & \multicolumn{2}{c}{\bf CLIP}\\
        \cmidrule{2-5}
         & \bf Acc (\%) $\uparrow$ & \bf KL  $\downarrow$ & \bf Acc (\%) $\uparrow$ & \bf KL $\downarrow$ \\
        \midrule
        TI (CKM) & $36.35\pm11.34$ &$0.1432\pm0.0637$ & $34.16\pm14.55$ & $0.1386\pm	0.0101$\\
         Ours & $\bm{51.25\pm4.11}$ & $\bm{0.0736\pm0.0626}$ & $\bm{45.94\pm2.94}$ & $\bm{0.0720\pm0.0969}$
        \\
        \bottomrule
    \end{tabular}}
    \vspace{-5pt}
    \captionof{table}{\small \textbf{Quantitative Evaluation on  $\bm{S_1}$ across $3$ seeds.}}
    \label{table:classification_rebuttal}
    \vspace{-12pt}
\end{table}

\subsection{Unsupervised Concept Discovery} 
\label{supp:concept_discovery}
\begin{figure*}[t]
\small
\centering
\includegraphics[width=0.99\linewidth]{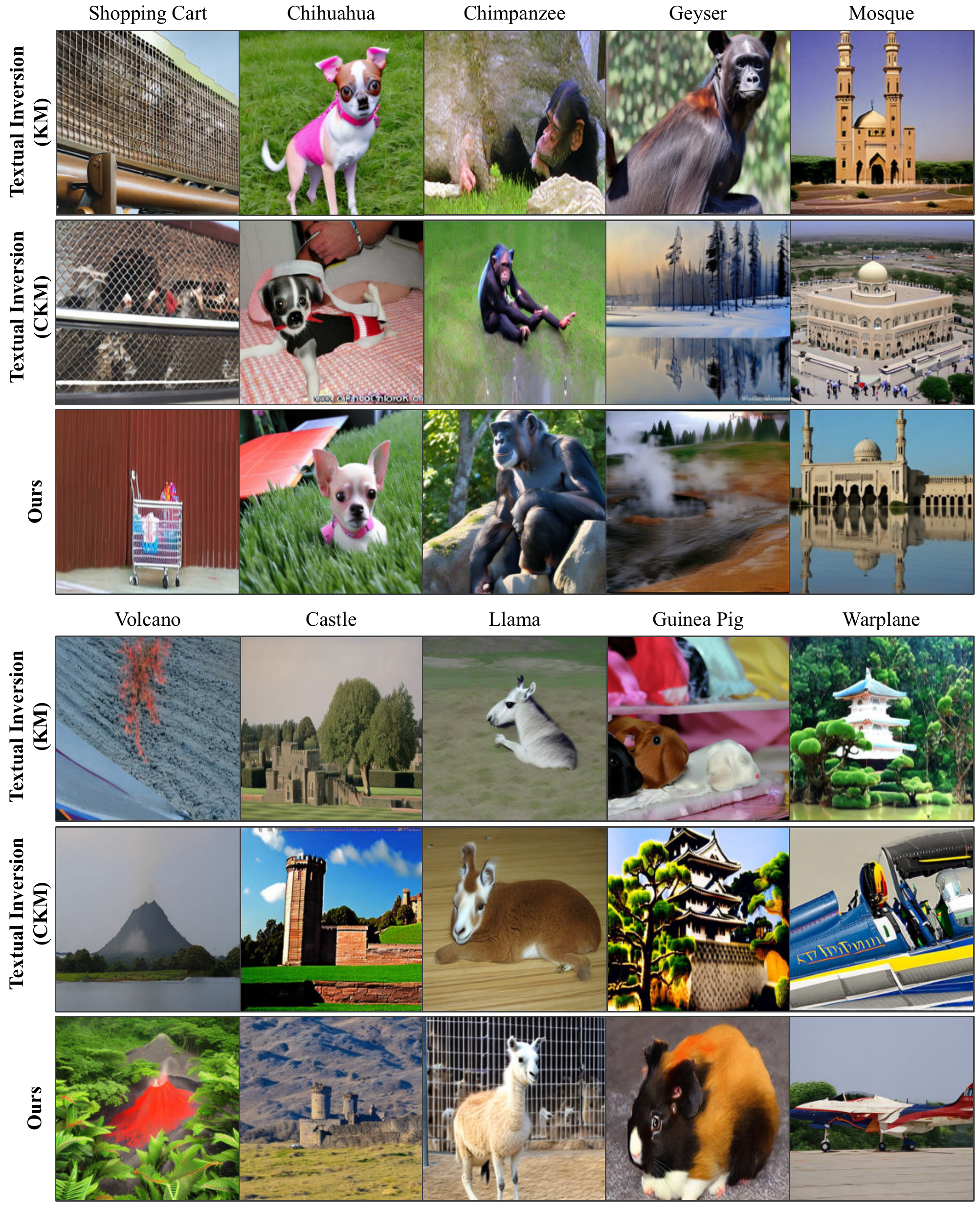}
\caption{
\textbf{Object Decomposition.} Object decomposition results on ImageNet $S_1$ (top) and $S_2$ (bottom). Note that concepts are labeled with our best interepretation for easy understanding.
}
\label{fig:imagenet_decomposition_supp_1}
\end{figure*}

\begin{figure*}[t]
\small
\centering
\includegraphics[width=0.99\linewidth]{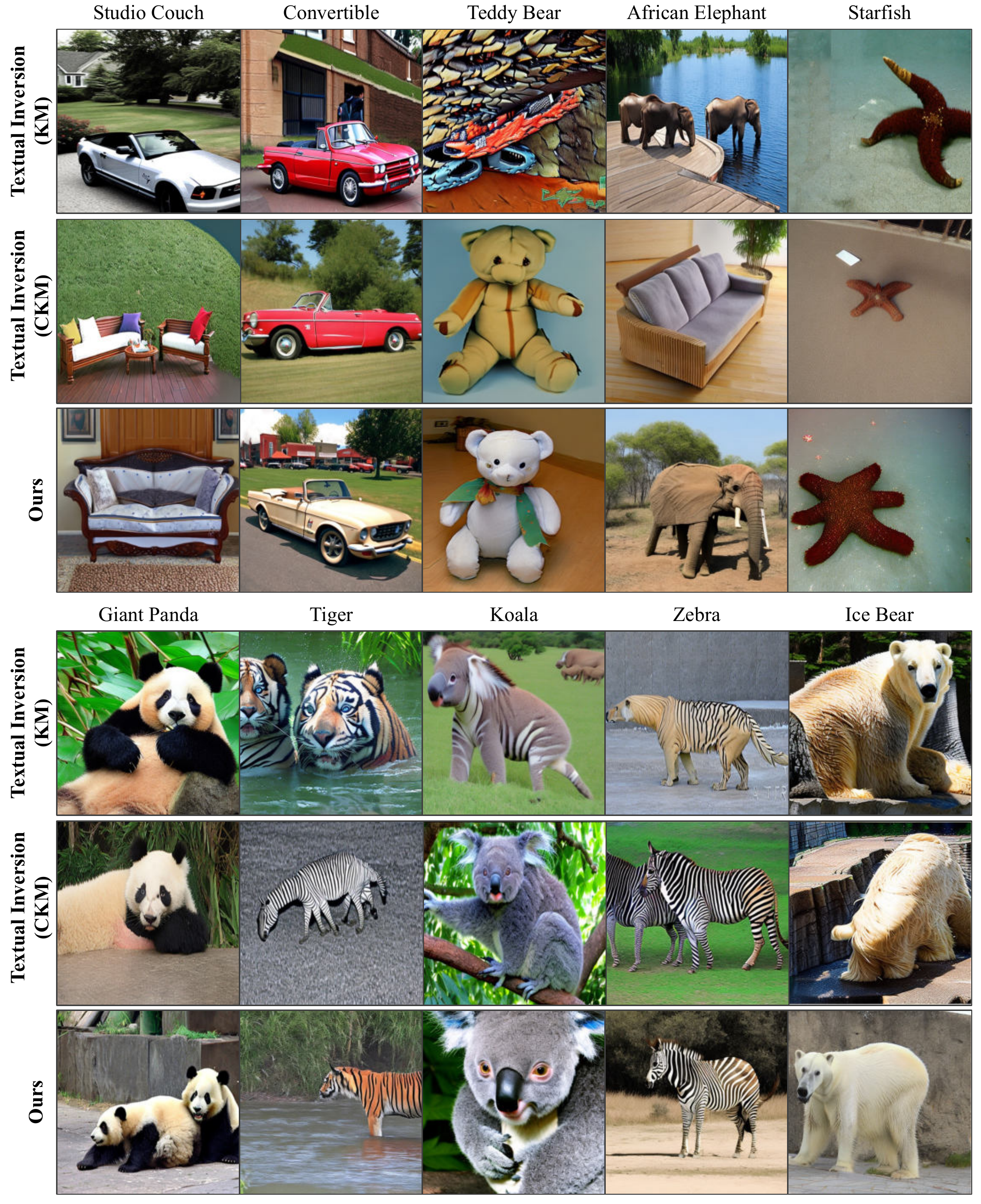}
\caption{
\textbf{Object Decomposition.} Object decomposition results on ImageNet $S_3$ (top) and $S_4$ (bottom). Note that concepts are labeled with our best interepretation for easy understanding.
}
\label{fig:imagenet_decomposition_supp_2}
\end{figure*}

\begin{figure*}[t]
\small
\centering
\includegraphics[width=1\linewidth]{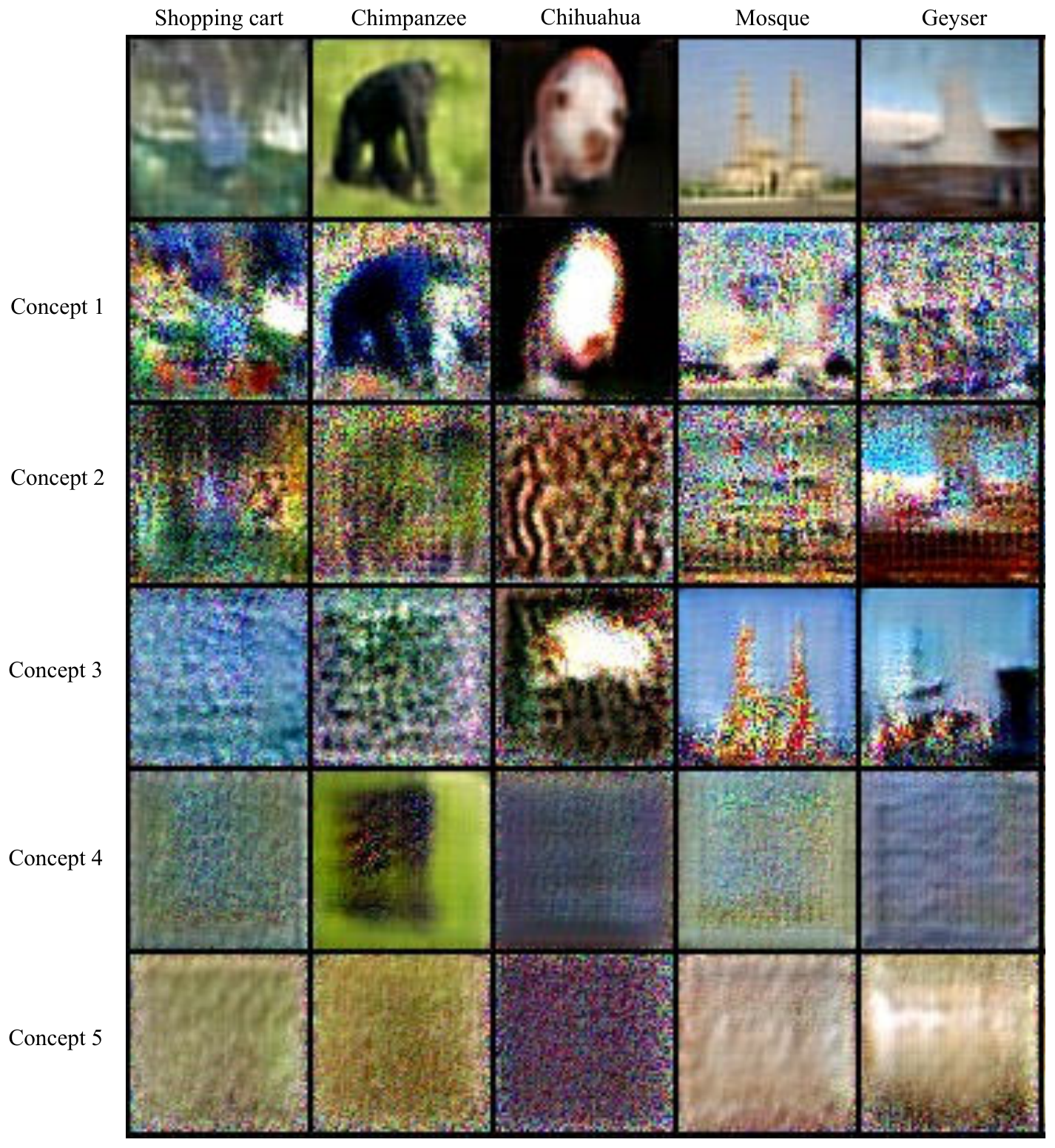}
\caption{
\textbf{Object Decomposition using COMET~\cite{du2021comet}.} Object decomposition results on ImageNet $S_1$, where $5$ of concepts learned from each training image (top) are not realistic.
}
\label{fig:imagenet_decomposition_comet}
\end{figure*}

\paragraph{Object Discovery.} We show qualitative comparisons between our method and baselines for each set of ImageNet experiments. We find that baselines generate repetitive concepts. For example, textual inversion (KM) discovers two embeddings for the class of chimpanzee, as shown in the \nth{3} and \nth{4} columns of the \nth{1} row in \fig{fig:imagenet_decomposition_supp_1}. Furthermore, both variants of textual inversion fail to generate certain concepts, such as \textit{shopping cart}, while our method can discover such concepts, as shown in the \nth{1} column of the \fig{fig:imagenet_decomposition_supp_1}. We demonstrate that such problems exist across all experiments in \fig{fig:imagenet_decomposition_supp_1} and \fig{fig:imagenet_decomposition_supp_2}.
We also train COMET on ImageNet to decompose images into object categories. However, it scales poorly to more complex images, thus failing to decompose such images into realistic concepts as illustrated in \fig{fig:imagenet_decomposition_comet}. 

\begin{figure*}[t]
\small
\centering
\includegraphics[width=1\linewidth]{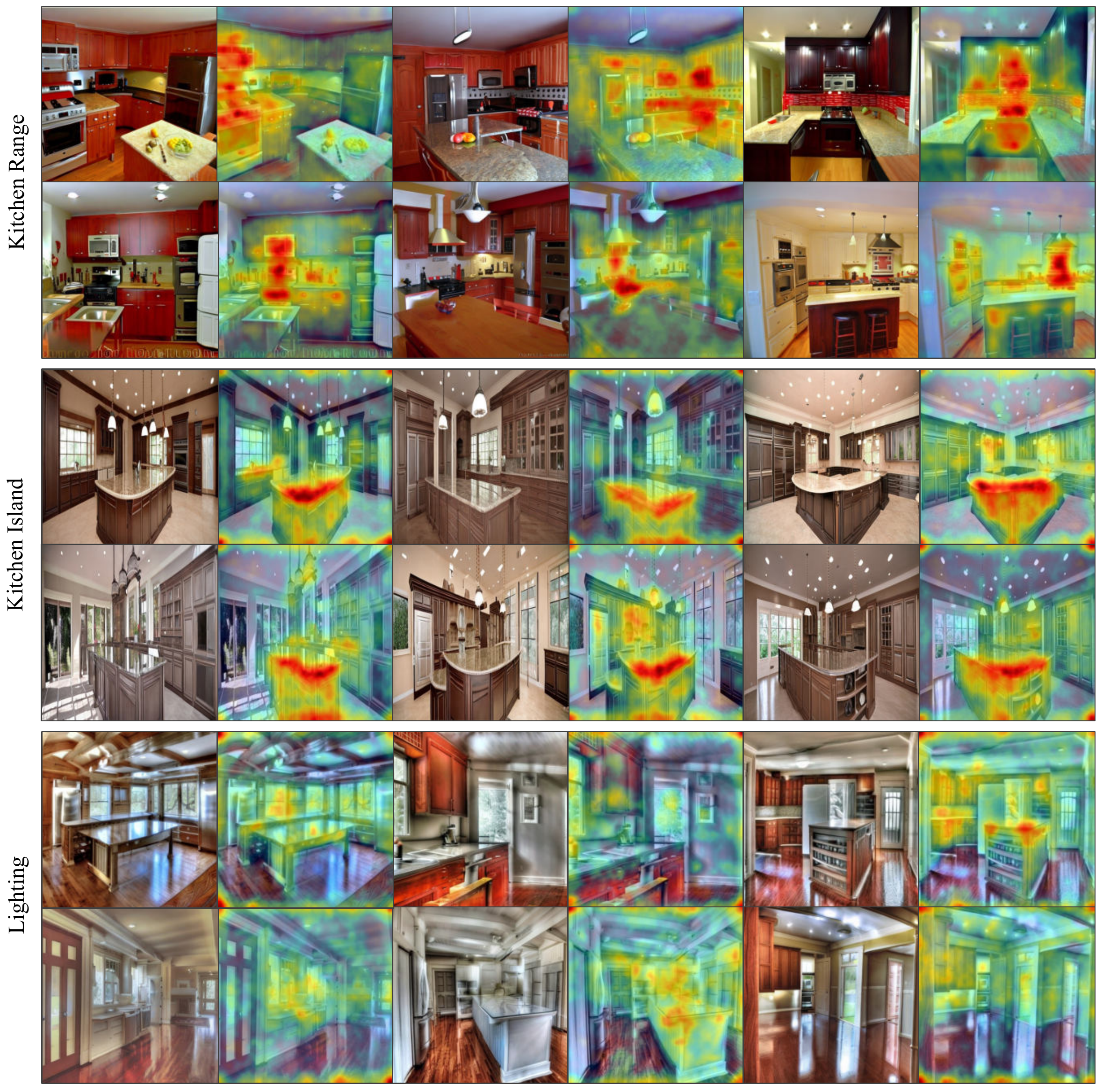}
\caption{
\textbf{Indoor Scene Decomposition.} We show additional results of decomposed kitchen concepts. Note that concepts are labeled with our best interepretation based on attention maps for easy understanding.
}
\label{fig:ade20k_decomposition_supp}
\end{figure*}

\paragraph{Indoor Scene Discovery.} 
To further verify effectiveness of our approach, we provide additional qualitative results of our method on indoor scene decomposition, specifically in the kitchen setting. In \fig{fig:ade20k_decomposition_supp}, we show both generated samples (odd columns) along their cross-attention maps (even columns) on three major concepts, incluing \textit{kitchen range}, \textit{kitchen islands}, and \textit{lighting effects}.

\begin{figure*}[t]
\small
\centering
\includegraphics[width=1\linewidth]
{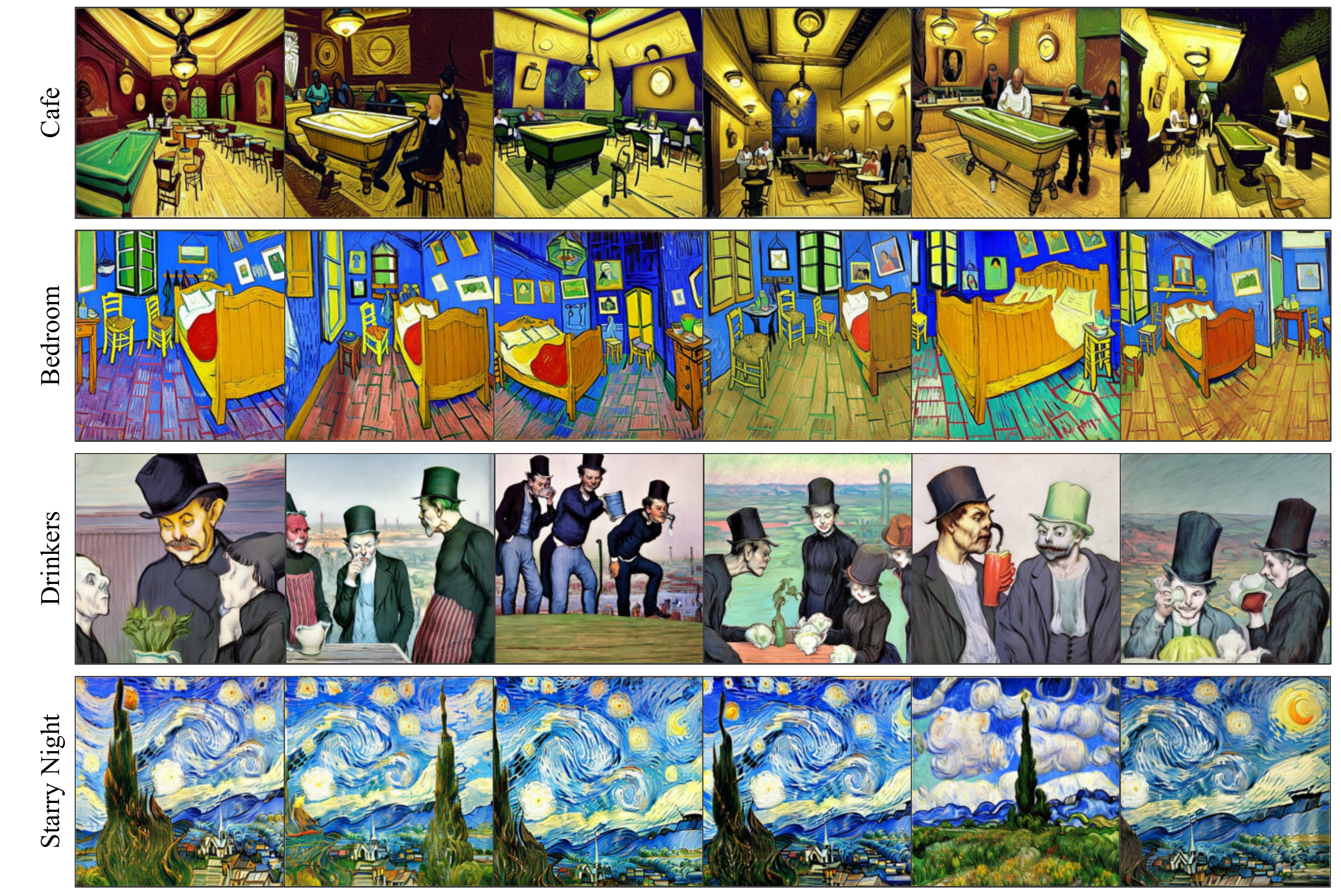}
\caption{
\textbf{Art Decomposition.} We show results of decomposed concepts using Van Gogh's paintings. Note that concepts are labeled with the name of the most similar paintings in the training set for easy understanding.
}
\label{fig:art_decomposition_supp_1}
\end{figure*}

\begin{figure*}[t]
\small
\centering
\includegraphics[width=1\linewidth]{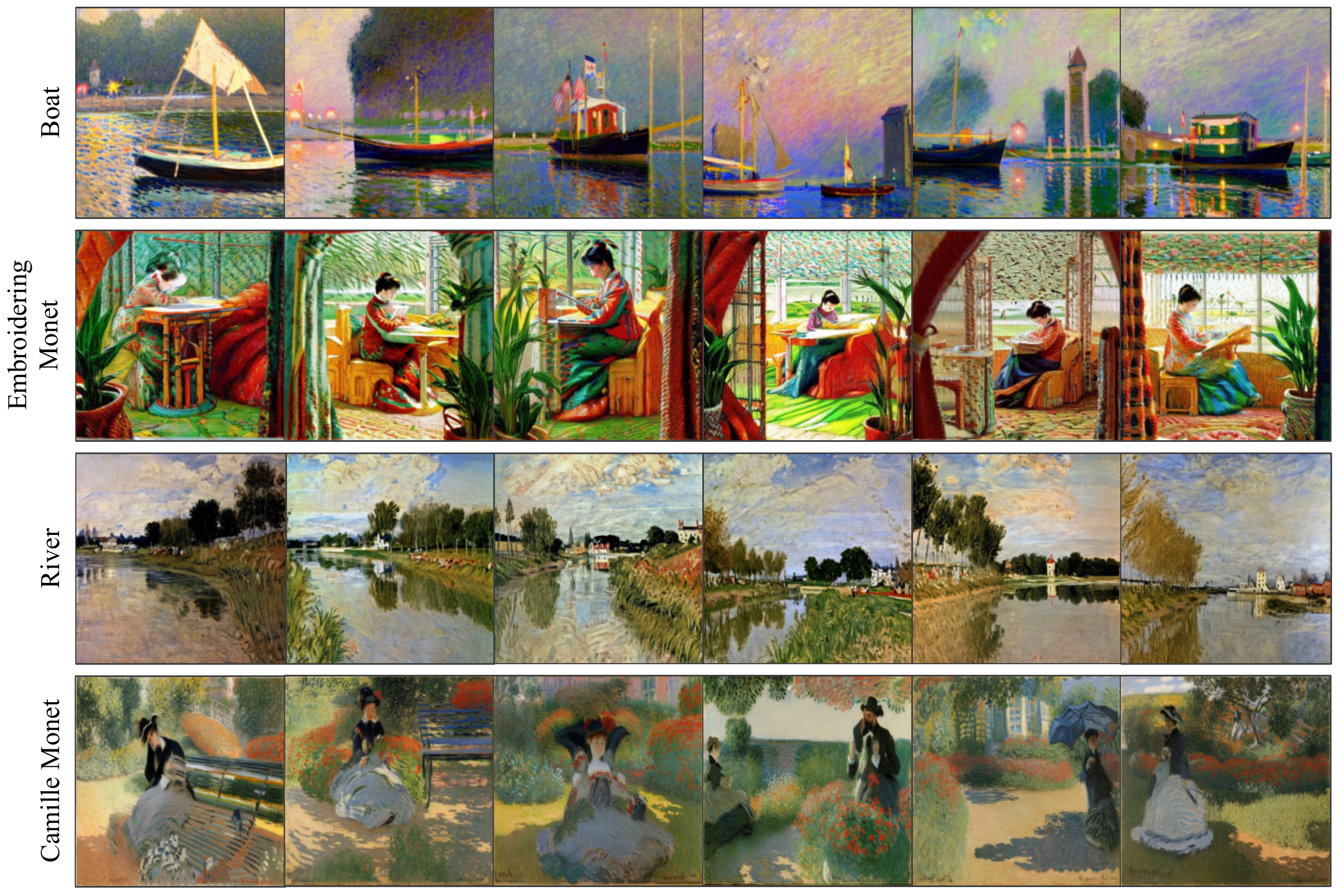}
\caption{
\textbf{Art Decomposition.} We show results of decomposed concepts using Claude Monet's paintings. Note that concepts are labeled with the name of the most similar paintings in the training set for easy understanding.
}
\label{fig:art_decomposition_supp_2}
\end{figure*}

\begin{figure*}[t]
\small
\centering
\includegraphics[width=1\linewidth]{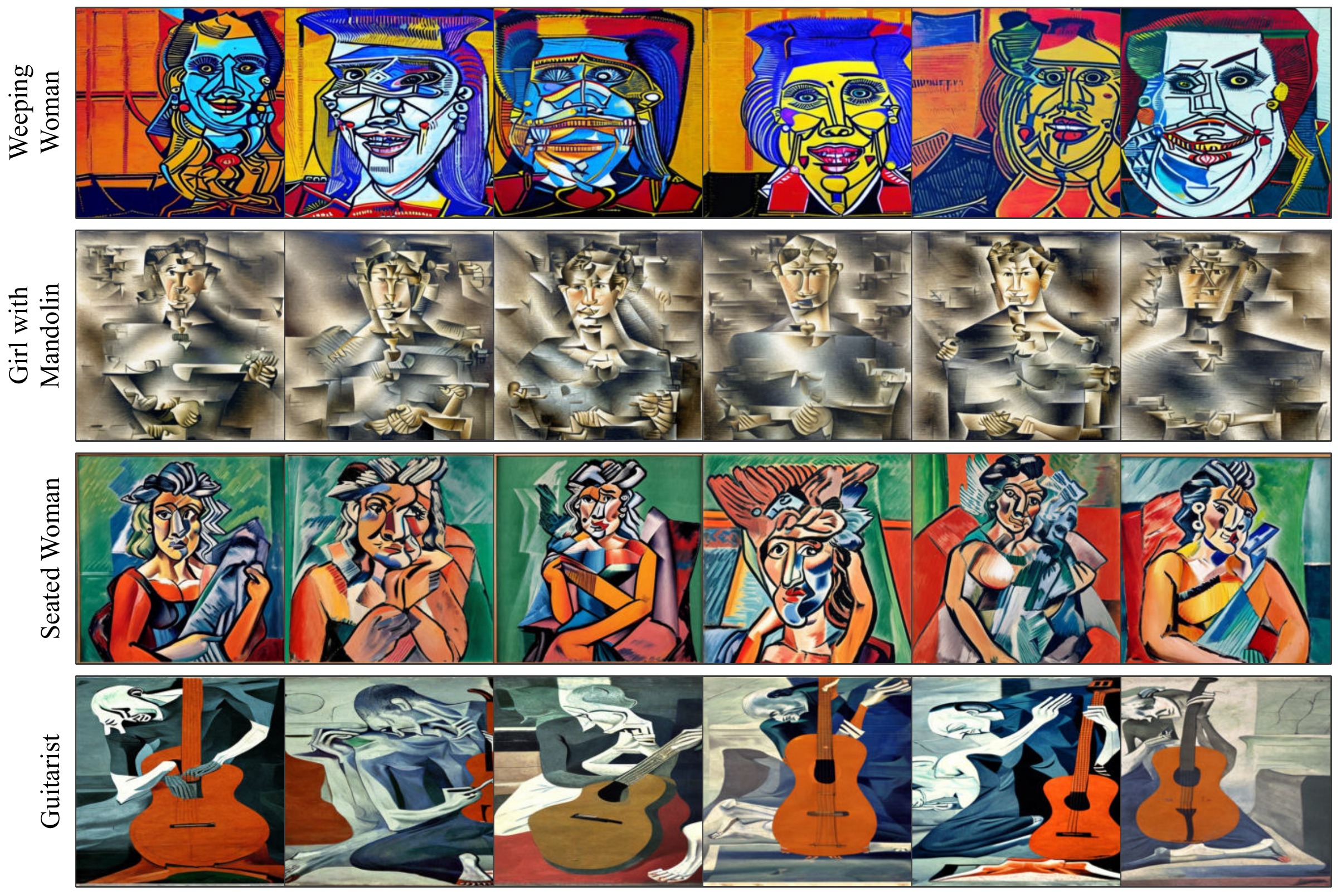}
\caption{
\textbf{Art Decomposition.} We show results of decomposed concepts using Pablo Picasso's paintings. Note that concepts are labeled with the name of the most similar paintings in the training set for easy understanding.
}
\label{fig:art_decomposition_supp_3}
\end{figure*}

\paragraph{Artistic Concept Discovery.} Finally, we show our decomposed concepts based on artistic paintings, including Van Gogh (\fig{fig:art_decomposition_supp_1}), Claude Monet (\fig{fig:art_decomposition_supp_2}) and Pablo Picasso (\fig{fig:art_decomposition_supp_3}). Our method can discover artistic concepts from few paintings. We provide names of original paintings on the leftmost side for easy understanding.

\begin{figure}[t]
\begin{center}
\includegraphics[width=0.47\textwidth]{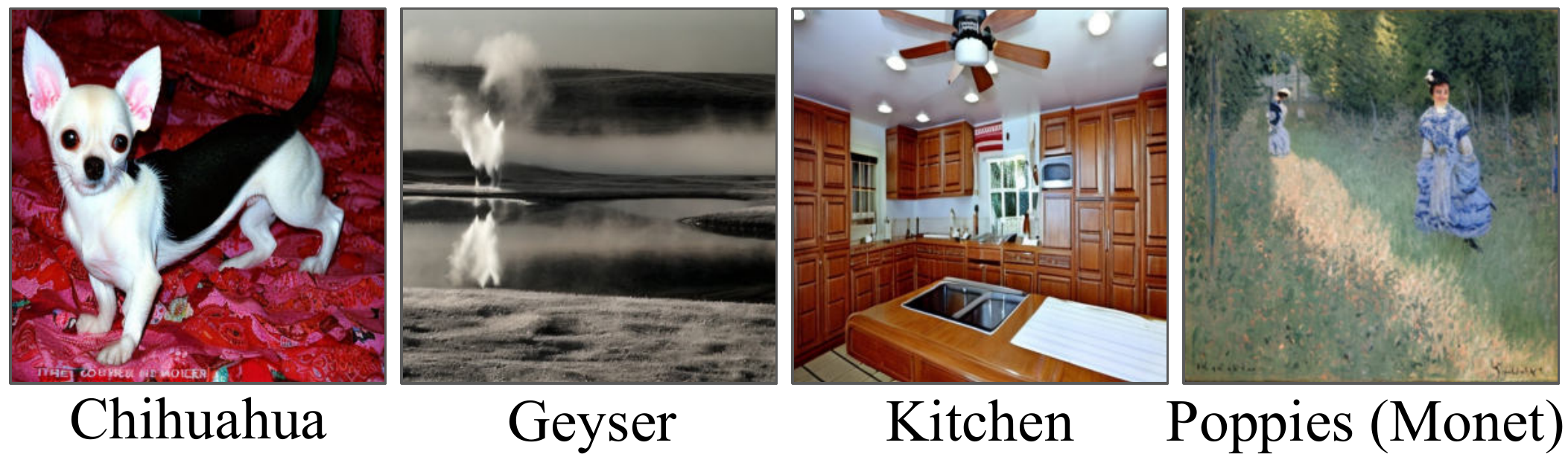}
\end{center}
\vspace{-15pt}
\caption{\small \textbf{Qualitative results on the hybrid dataset.}
}
\label{fig:hybrid_decomposition_rebuttal}
\end{figure}

\paragraph{Concept Discovery from hybrid modalities.} We run our method on a hybrid dataset that contains images from four concepts, \ie, kitchen, Geyser, Chihuahua, and Claude Monet paintings. As shown in Figure~\ref{fig:hybrid_decomposition_rebuttal}, our method can successfully discover all four distinct concepts.

\begin{table}[t]
    \centering
    \setlength{\tabcolsep}{3mm}
    \scalebox{0.80}{
    \begin{tabular}{c|c|c}
        \toprule
        \bf Dataset & \bf CLIP (Training Data) & \bf CLIP (Generation) \\
        \cmidrule{1-3}
        ADE20K & 0.1760 & 0.1701 \\
        Van Gogh & 0.1411 & 0.1259  \\
        ImageNet $S_1$ & 0.1089 & 0.1188 \\
        \bottomrule
    \end{tabular}}
    \vspace{-5pt}
    \captionof{table}{\small \textbf{Quantitative Evaluation on Image Diversity.} }
    \label{table:clip_rebuttal}
\end{table}

\subsection{Composing Discovered Concepts}
\label{supp:composing_concepts}
\begin{figure*}[t]
\small
\centering
\includegraphics[width=1\linewidth]{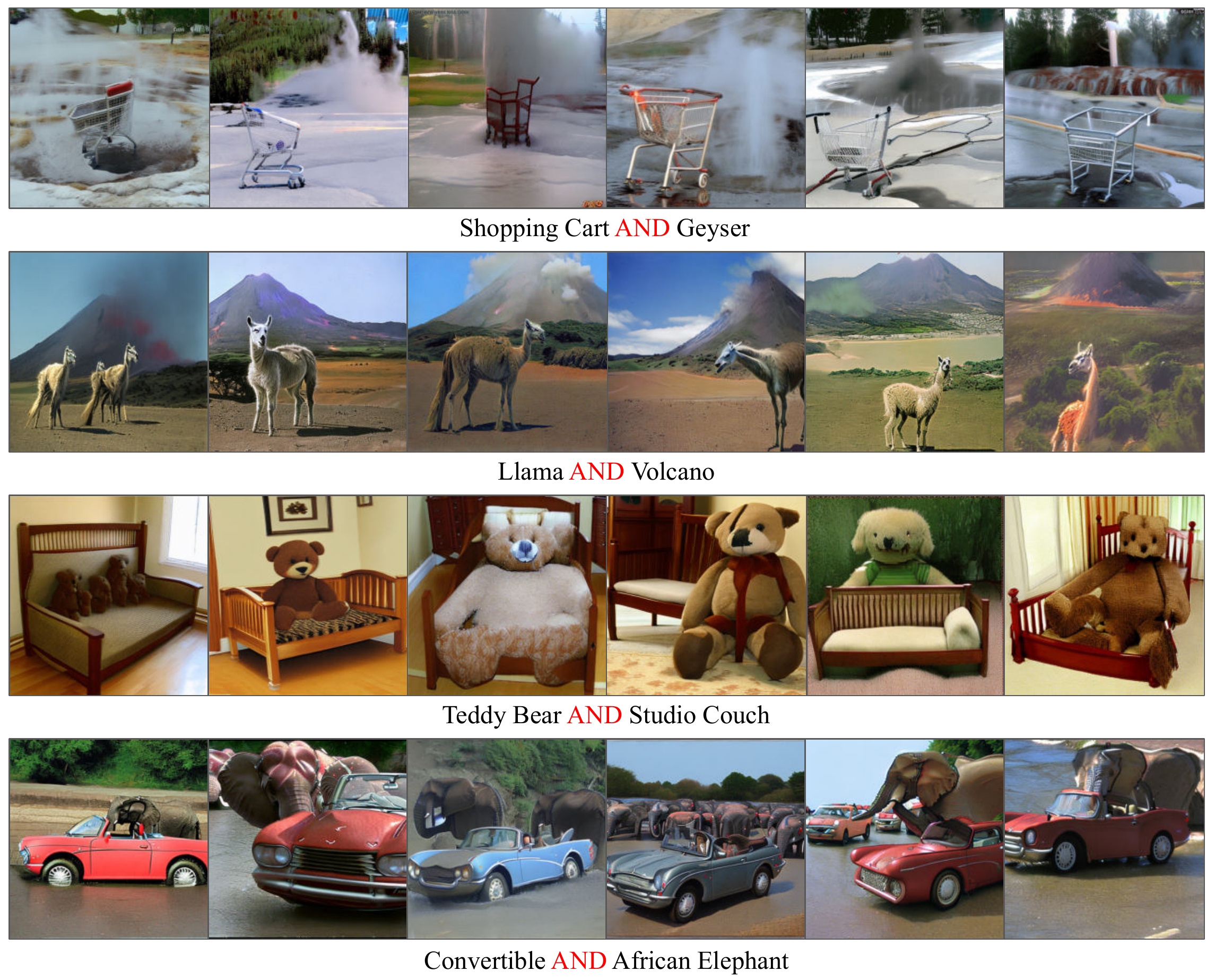}
\caption{
\textbf{Object Composition.} We show additional results of object composition using ImageNet classes. Note that concepts are labeled with our best interepretation for easy understanding.
}
\label{fig:imagenet_composition_supp}
\end{figure*}

\paragraph{Object Composition.} We show our method enables multi-object composition in \fig{fig:imagenet_composition_supp}. For example, we can generate images that resemble ``a teddy bear sitting on a studio couch'' in the \nth{3} row by composing two discovered classes.

\begin{figure*}[t]
\small
\centering
\includegraphics[width=1\linewidth]{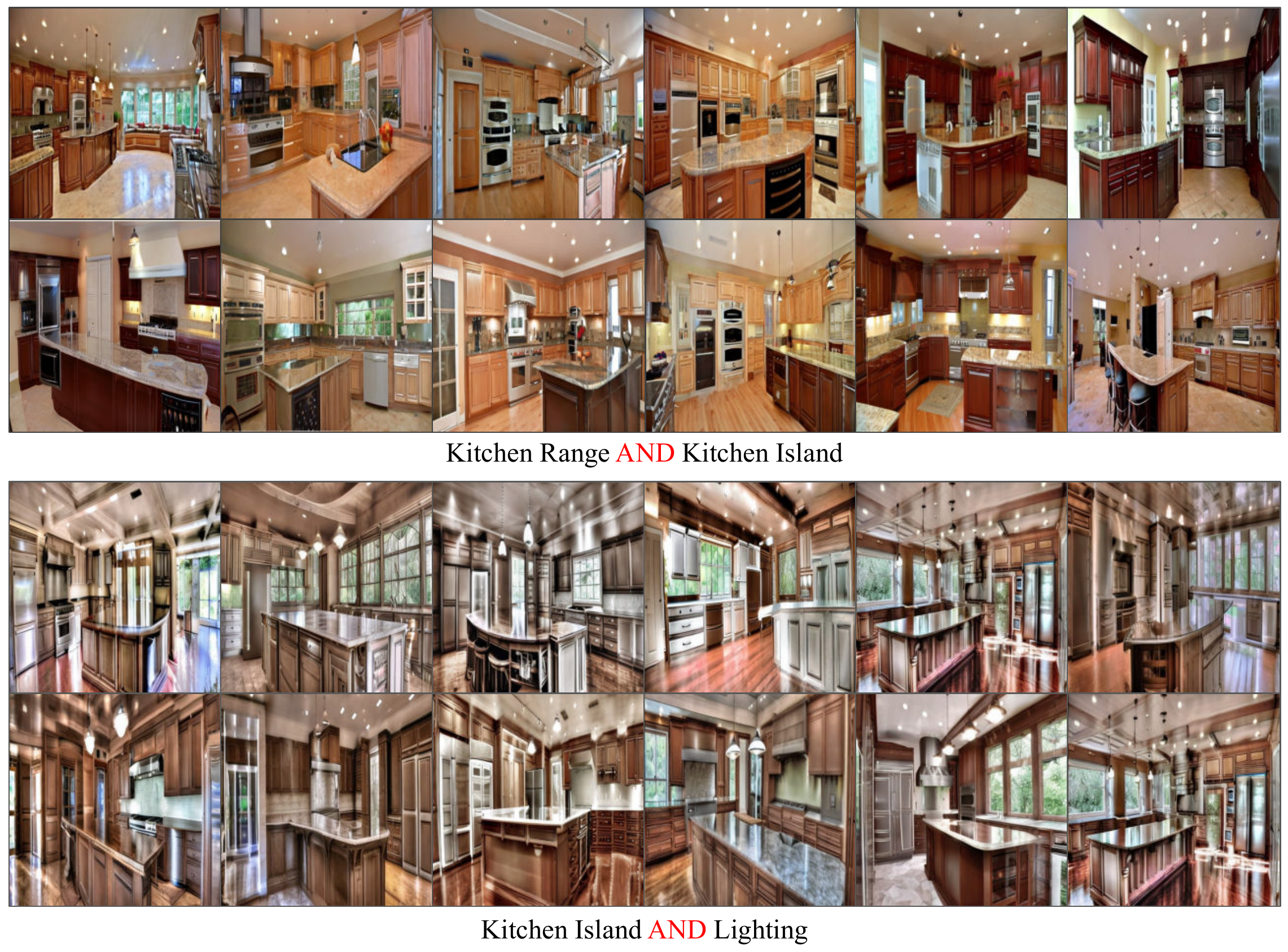}
\caption{
\textbf{Kitchen Scene Composition.} We demonstrate results of composing discovered kitchen components. Note that concepts are labeled with our best interpretation of what they are for easy understanding.
}
\label{fig:ade20k_composition_supp}
\end{figure*}

\paragraph{Scene Composition.} In \fig{fig:ade20k_composition_supp}, we further compose indoor kitchen components to generate indoor scenes that contain given specifications, including combinations of \textit{kitchen range} and \textit{lighting effects} in \nth{2} row.

\begin{figure*}[t]
\small
\centering
\includegraphics[width=1\linewidth]{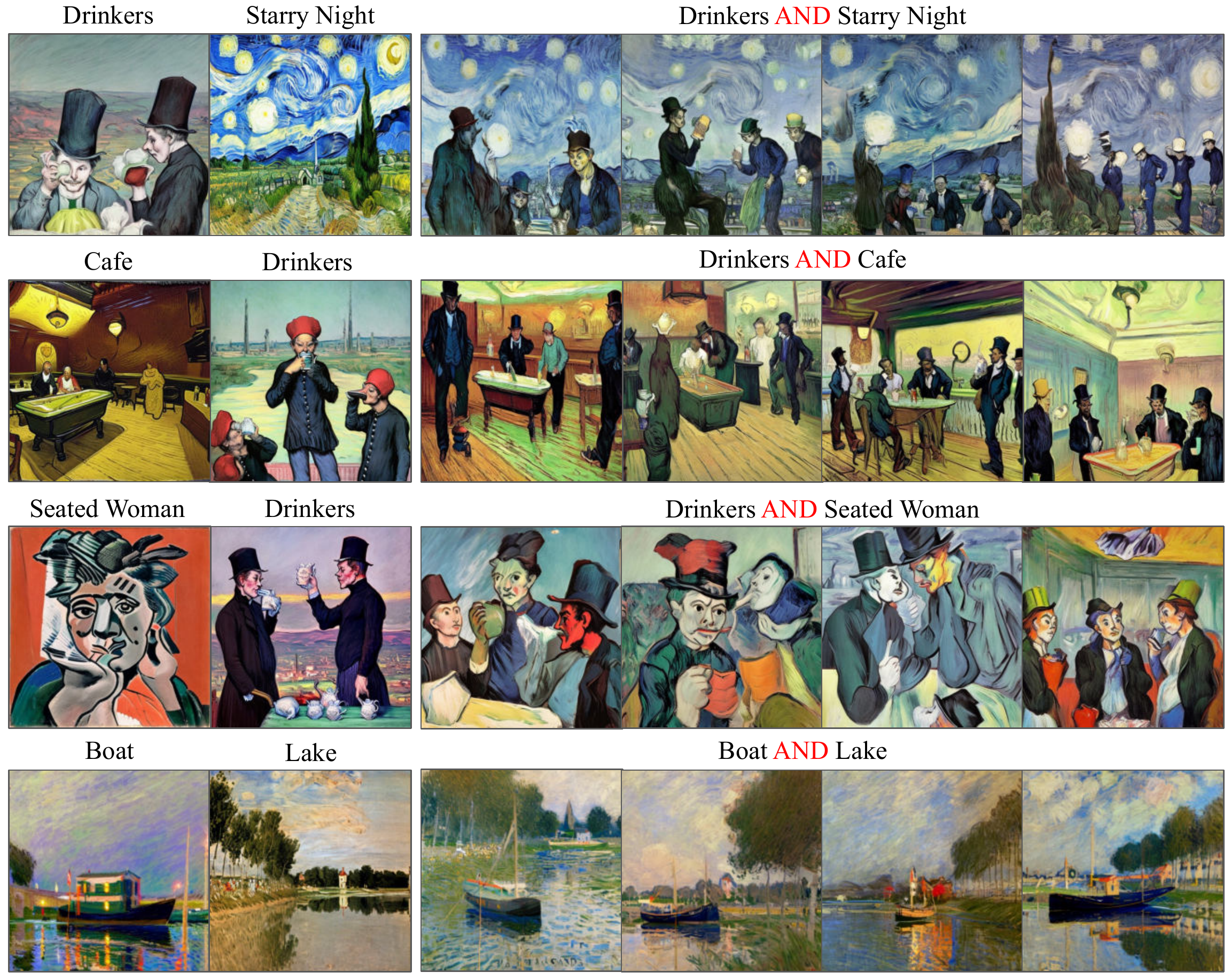}
\caption{
\textbf{Style Composition.} We show composition of different concepts discovered from differnet paintings. Note that concepts are labeled with the names of the most similar paintings in the training set.
}
\label{fig:art_composition_supp}
\end{figure*}

\paragraph{Style Composition.} We also demonstrate compositioanl results of decomposed concepts from artistic paintings in \fig{fig:art_composition_supp}. In this experiment, we either compose concepts discovered from the same artistic (\eg, \nth{1} row) or even combine concepts across different artists (\eg, \nth{3} row).

\begin{figure*}[t]
\small
\centering
\includegraphics[width=1\linewidth]{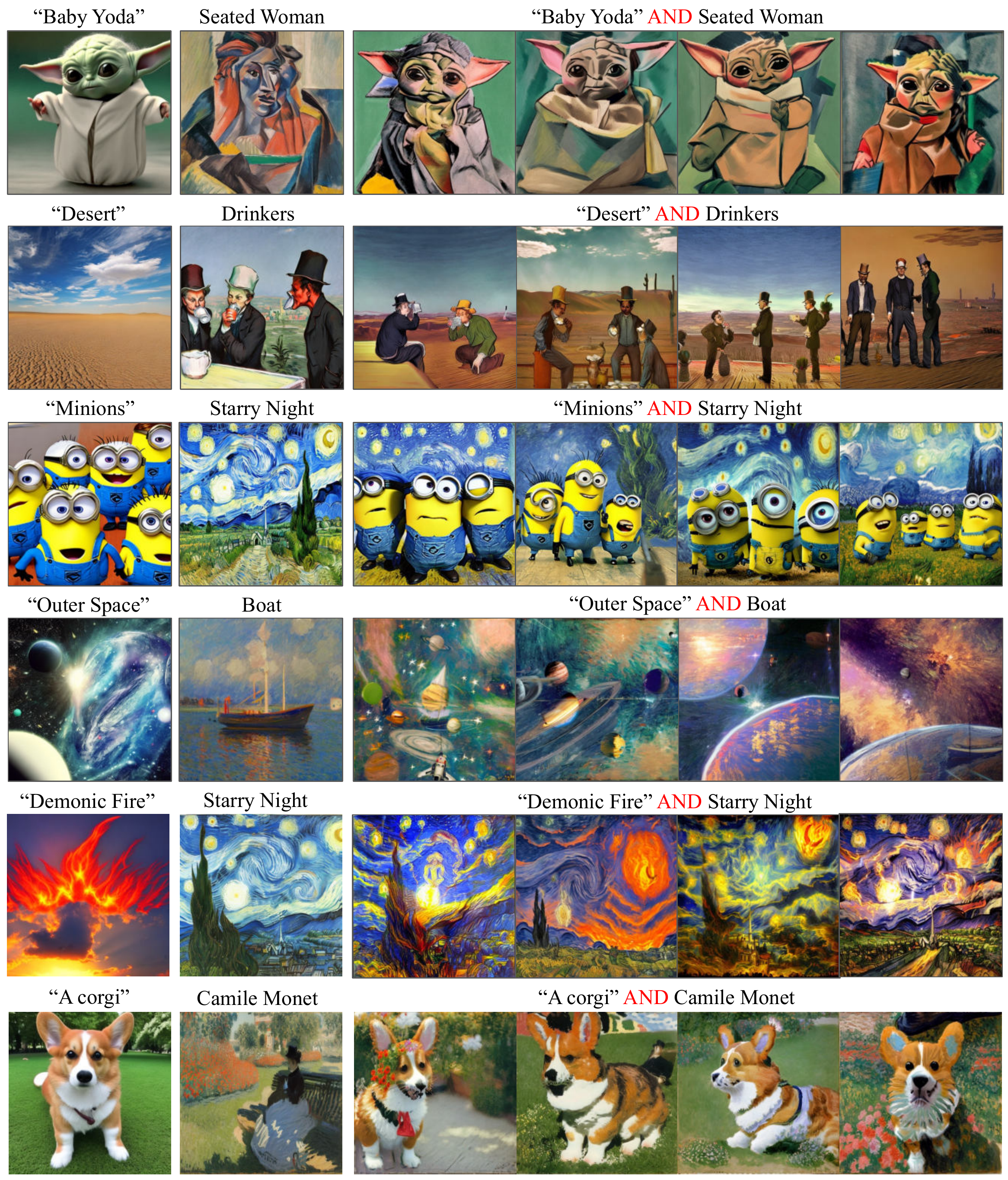}
\caption{
\textbf{External Composition.} We show composition results (\nth{3} column) of existing concepts (\nth{1} column) and discovered concepts (\nth{2} column), where discovered concepts are labeled with the names of the most similar paintings in the training set for easy understanding.
}
\label{fig:external_composition_supp}
\end{figure*}

\paragraph{External Composition.} Finally, we provide additional results of external composition, where we compose existing concepts (\eg, text) with discovered concepts in \fig{fig:external_composition_supp}. We show that we can enable style transfer by composing  text descriptions shown in the \nth{1} column and discovered concept in the \nth{2} column to generate images.

\begin{figure}[t]
\small
\begin{center}
\includegraphics[width=0.45\textwidth]{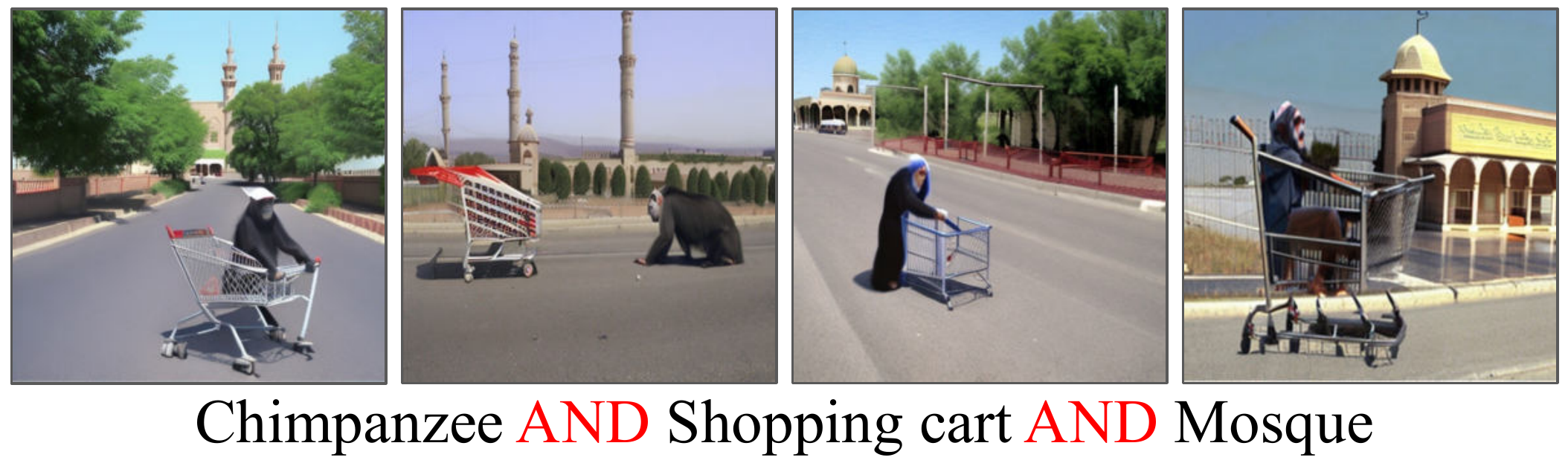}
\end{center}
\vspace{-15pt}
\caption{\small \textbf{Object Composition of $3$ discovered concepts.}}
\label{fig:composition_rebuttal}
\end{figure}

\paragraph{Composition of multiple concepts.} 
Our method can compose more than $2$ concepts.
In Figure~\ref{fig:composition_rebuttal}, we show the composition of $3$ concepts discovered from ImageNet $S_1$.

\section{Details of Datasets}
\label{supp:datasets}
\paragraph{ImageNet~\cite{deng2009imagenet}.} We use $4$ sets of ImageNet class combinations, denoted as ImageNet $S_1$, $S_2$, $S_3$ and $S_4$ in our experiments. Each combination consists of $5$ object categories. $S_1$ includes \textit{geyser}, \textit{Chihuahua}, \textit{chimpanzee}, \textit{shopping cart} and \textit{mosque}. $S_2$ includes \textit{guinea pig}, \textit{warplane}, \textit{castle}, \textit{llama} and \textit{volcano}.
$S_3$ includes \textit{convertible}, \textit{starfish}, \textit{studio couch}, \textit{african elephant} and \textit{teddy}.
$S_4$ includes \textit{koala}, \textit{ice bear}, \textit{zebra}, \textit{tiger} and \textit{giant panda}. We randomly choose $5$ images per category for each set as our training data.

\paragraph{ADE20K~\cite{zhou2017scene}.} We use images in the \textit{bedroom} subcatergory under the category of \textit{home or hotel} for our training. Similarly, we randomly select $25$ images as our training dataset.

\section{Details of Baselines}
\label{supp:baselines}
\paragraph{COMET~\cite{du2021comet}.} Most relevant to our work, COMET uses a set of EBMs to discover concepts in an unsupervised manner. However, COMET decomposes each individual image into a set of concepts, while our method decomposes a set of images into a set of concepts. Hence, COMET doesn't enable novel generation of the decomposed concepts and we instead visualize decomposed components from training images. For training, we use $5$ components representing $5$ object categories to train COMET using the default training setting from the official codebase. 

\paragraph{Textual Inversion~\cite{gal2022image}.} Given a set of similar images, textual inversion optimizes a single concept $c$, thus assuming a correspondence between the training data and the concept. In our experiments, however, we train an unconditional textual inversion by optimizing one single concept using all training images regardless of image classes or concepts. During inference, we generate $320$ images using the prompt: ``a photo of $c$'' for evaluation.

\paragraph{Textual Inversion (KM).} In this paper, our goal is to discover multiple concepts in an unsupervised way. Thus, we utilize unsupervised algorithms, such as K-means clustering, to obtain pseudo-labels. Before training a textual inversion model, we run K-means on the training images in pixel space to obtain predicted labels, which are used to optimize corresponding concepts during training. In our experiments, each ImageNet set has training images from $5$ categories, so we initialize $5$ concepts for optimization. During inference, we sample $64$ images per concept for evaluation.
 
\paragraph{Textual Inversion (CKM).} We also use another variant of textual inversion and K-means clustering as our baseline. In this case, we run K-means on image latent representations encoded by CLIP~\cite{radford2021learning}, thus we name it as CLIP-based K-means (\ie, CKM for brevity). Similarly, we evaluate this baseline in the same way as textual inversion (KM).

\paragraph{Training Details.} We train every single model with a batch size of $2$ and $8$ gradient accumulation steps, and thus an effective batch size of $16$ per iteration for a total number of $3000$ iterations on each dataset using a single NVIDIA A40 GPU. Other hyper-parameters (\eg, optimizer) are the same as the original textual inversion codebase~\cite{gal2022image}.

\section{Details of Our Approach}
\label{supp:our_approach}
\paragraph{Training.} 
To discover compositional concepts we our approach, we initialize $M$ (\ie, $5$) words along with their random embeddings as our concepts in our experiments, and a weight matrix with a shape of $N \times M$, where $N$ is the number of training images. Then we utilize our method to optimize both weights and all $M$ embeddings for each training image, as shown in \fig{fig:model}. Training details are the same as that of baselines shown in \cref{supp:baselines}, where embeddings are optimized with a batch of 16 for 3000 iterations.

\paragraph{Inference.} To enable image generation of each discovered concept, we sample images using each word using classifier-free guidance~\cite{ho2022classifier}. For compositional generation, we sample images using conjunction operator (\ie, AND) from composable diffusion~\cite{liu2022compositional}.

\end{document}